\documentclass{article} 
\usepackage{iclr2027_conference,times}

\usepackage{amsmath,amsfonts,bm}

\def\eqref#1{equation~\ref{#1}}

\def\1{\bm{1}}

\DeclareMathAlphabet{\mathsfit}{\encodingdefault}{\sfdefault}{m}{sl}
\SetMathAlphabet{\mathsfit}{bold}{\encodingdefault}{\sfdefault}{bx}{n}

\usepackage{url}
\usepackage{algorithm}
\usepackage{algpseudocode}
\usepackage{amsmath}
\usepackage{booktabs}
\usepackage{graphicx}
\usepackage{placeins}
\usepackage{flafter}
\usepackage{hyperref}
\usepackage[table]{xcolor} 
\usepackage{arydshln}      
\usepackage{multirow}
\usepackage{booktabs}
\usepackage{subcaption}
\usepackage{adjustbox}
\usepackage{afterpage}
\usepackage{listings}
\newsavebox{\ipTwoPLeft}
\newsavebox{\unseenLeft}
\usepackage{array}
\newcolumntype{N}{>{\fontsize{7}{8.5}\selectfont}c}

\title{SPIN: Image Immunization Against Diffusion Editing via Single-Step Projection in Stochastic Neighborhoods}

\author{
Fengming Gu$^{1,2,3}$, 
Jie Zhang$^{2,3}$, 
Zhongqi Wang$^{2,3}$, 
Qiankun Li$^{4}$, \\
\textbf{Shiguang Shan$^{2,3}$, 
Xilin Chen$^{2,3}$} \\
$^{1}$School of Advanced Interdisciplinary Sciences, 
University of Chinese Academy of Sciences \\
$^{2}$State Key Laboratory of AI Safety, 
Institute of Computing Technology, Chinese Academy of Sciences \\
$^{3}$University of Chinese Academy of Sciences, $^{4}$Imperial College London \\
\texttt{gufengming18@mails.ucas.ac.cn}, 
\texttt{q.li2@imperial.ac.uk} \\
\texttt{\{zhangjie, wangzhongqi23s, sgshan, xlchen\}@ict.ac.cn}
}

\newcommand{\paneldash}{%
  \hspace{2.5pt}\rule[-0.10em]{0.35pt}{0.45em}\hspace{2.5pt}}

\iclrfinalcopy 
\begin{document}
\addtocontents{toc}{\protect\setcounter{tocdepth}{-1}}

\maketitle

\begin{abstract}

Diffusion models have greatly advanced instruction-guided image editing, while also raising concerns about unauthorized image manipulation. Image immunization addresses this risk by adding imperceptible perturbations to an input image to disrupt subsequent edits. Since editing requests are unknown at image release, protection should remain effective beyond the instruction used to construct the perturbation. Existing immunization methods either require costly full-trajectory backpropagation or use intermediate objectives whose effects may be weakened by subsequent denoising. Meanwhile, a single inference path provides limited feedback about alternative denoising continuations. To address these challenges, we propose \textsc{SPIN}, a framework for image immunization via one-step projection over local stochastic trajectory neighborhoods. Starting from an early denoising state, \textsc{SPIN} generates stochastic neighboring states under the same instruction and predicts their clean latents through one-step projection without full unrolling. We then optimize a bounded input perturbation to maximize the average deviation of these predictions from a clean-edit reference, encouraging the perturbation to disrupt multiple possible editing outcomes. Experiments on two image editors demonstrate substantial gains in protection performance, with \textsc{SPIN} outperforming compared methods across all six metrics under seen instructions and in the more challenging unseen instruction setting.
\end{abstract}

\section{Introduction}

Instruction-guided diffusion models have made it possible to modify a source image using natural-language instructions while preserving content unrelated to the requested changes~\citep{brooks2023instructpix2pix,zhao2024ultraedit}. Publicly available image editing models have substantially lowered the barriers to personalized image manipulation. However, the same accessibility also makes publicly released images vulnerable to unauthorized editing, which may compromise personal privacy and reputation or facilitate misleading content. Deepfake detection~\citep{rossler2019faceforensics++} and AI-generated image detection~\citep{ojha2023towards} can identify manipulated content, but operate only after such content has been produced. This motivates proactive protection that is embedded imperceptibly in images before release.

Image immunization protects source images by adding imperceptible perturbations that disrupt subsequent unauthorized edits. A direct approach optimizes perturbations against the final editing output, but backpropagation through the denoising trajectory incurs substantial computational cost~\citep{salman2023raising}. More efficient alternatives construct surrogate objectives from denoising predictions or intermediate representations~\citep{liang2023advdm,liang2023mist}. Such objectives can target encoder representations~\citep{xue2024sds}, instruction-related cross-attention~\citep{lo2024distraction}, or early denoising states~\citep{choi2025diffusionguard}, providing protection signals through shorter gradient paths.

Nevertheless, deviations induced by intermediate representations may be attenuated by subsequent denoising, weakening their effect on the final edit. Protection under a fixed editing instruction may also depend on the specific sampling trajectory. This dependence becomes particularly relevant when future editing requests are unknown, as a single trajectory offers limited feedback about other possible continuations. The challenge is therefore to construct an effective optimization signal that reflects the perturbation's impact on editing outcomes across multiple possible continuations, enabling protection beyond the original editing instruction.

We address this challenge by proposing \textsc{SPIN} (Single-step Projection in Stochastic Neighborhoods), an image-immunization framework that uses predicted editing outcomes to guide protective perturbations. Single-step projection predicts a clean latent from an early denoising state. The discrepancy between this prediction and a fixed clean-edit reference provides an effective surrogate for optimizing perturbations that disrupt the final editing output. To extend this signal beyond a single inference path, \textsc{SPIN} generates neighboring candidate states through stochastic transitions from a shared early state. We then optimize a bounded input perturbation to maximize the average discrepancy between their projected outcomes and the reference, reducing dependence on a single construction path and promoting generalization to unseen editing instructions.

Experiments on SD3 and InstructPix2Pix show that \textsc{SPIN} outperforms all  baselines across six metrics under both seen and unseen editing instructions. Immunization success rates reach 90\% and 94\% under seen instructions (85\% and 91\% under unseen instructions), consistently exceeding those of competing methods. Ablations demonstrate that single-step projection provides an effective protection signal, while stochastic neighborhood sampling further improves generalization over single-candidate projection.

Our main contributions are summarized as follows:
\vspace{-4pt}
\begin{itemize}
    \item We propose \textsc{SPIN}, an image-immunization framework that uses single-step projection to guide perturbation optimization with predicted editing outcomes, providing an effective signal for disrupting the final edit.
\vspace{-3pt}
    \item We introduce stochastic neighborhood optimization to construct a single perturbation that targets multiple possible continuations of the editing process. By driving their projected outcomes away from the clean-edit reference, this strategy promotes disruption across different inference paths and improves generalization to unseen editing instructions.
\vspace{-3pt}
    \item Experiments under standard editing pipelines show that \textsc{SPIN} consistently outperforms the evaluated baselines in immunization success rate under both seen and unseen instructions, across flow-matching and noise-prediction diffusion editors.
\end{itemize}

\section{Related Work}

\subsection{Instruction-Guided Diffusion Image Editing}

Diffusion models support image editing by combining a generative prior with a source image and user-specified control. SDEdit~\citep{meng2022sdedit} perturbs and denoises an input through a stochastic differential equation to balance faithfulness and realism. Later methods introduced more explicit semantic control. Prompt-to-Prompt~\citep{hertz2023prompt} manipulates cross-attention maps to preserve spatial layout while changing textual content, whereas Imagic~\citep{kawar2023imagic} optimizes a text embedding and fine-tunes the model for semantic edits on a single image. InstructPix2Pix~\citep{brooks2023instructpix2pix} establishes a supervised learning paradigm for directly following natural-language editing instructions without per-image inversion or fine-tuning. UltraEdit~\citep{zhao2024ultraedit} advances this paradigm through large-scale editing supervision with diverse instructions, real-image anchors, and region annotations, supporting the adaptation of pretrained generative models, including Stable Diffusion 3~\citep{esser2024scaling}, to instruction-guided image editing. These advances improve the capability and accessibility of image editing, while also exposing publicly released images to increasingly diverse unauthorized transformations.

\subsection{Adversarial Examples}
Norm-bounded adversarial perturbations can alter model predictions~\citep{goodfellow2015adversarial}, and projected-gradient methods~\citep{madry2018pgd} formulate their construction as constrained optimization. To improve cross-model transferability, MI-FGSM~\citep{dong2018boosting} stabilizes updates through momentum, while VMI-FGSM~\citep{wang2021enhancing} further refines them using neighborhood gradient information. EOT~\citep{athalye2018eot} averages attack objectives over input transformations to improve transformation robustness, whereas universal adversarial perturbations generalize across images~\citep{moosavi2017universal}. SPIN targets the generation trajectories of diffusion editors, optimizing a bounded input perturbation over local stochastic continuations to disrupt generation and drive final editing outputs away from their clean counterparts.

\subsection{Image Immunization Against Diffusion Editing}
Image immunization methods mainly differ in the optimization signal used to construct protective perturbations. PhotoGuard exposes the central trade-off through an efficient encoder-targeting objective (PhotoGuard-E) and a more output-related but expensive full-diffusion objective (PhotoGuard-D)~\citep{salman2023raising}. AdvDM optimizes the diffusion denoising loss, while MIST combines semantic and textural objectives to disrupt the generation process at multiple representation levels~\citep{liang2023advdm,liang2023mist}. Subsequent methods shorten the gradient path by targeting specific internal components: SDS~\citep{xue2024sds} applies score distillation at the vulnerable VAE bottleneck, EditShield~\citep{chen2024editshield} shifts instruction-conditioned latent representations, and SA~\citep{lo2024distraction} disrupts prompt-relevant cross-attention features. DiffusionGuard~\citep{choi2025diffusionguard} places its objective at early denoising stages. Other approaches improve complementary aspects of protection: DCT-Shield~\citep{bala2025dctshield} optimizes frequency-domain perturbations for visual fidelity and JPEG resistance, whereas SIFM~\citep{dong2025sifm} combines intermediate-feature manipulation with semantic and perceptual criteria for evaluating edit failure. Despite their different targets, these methods either require costly differentiation through the diffusion trajectory or rely on internal surrogates whose deviations may be attenuated during subsequent denoising.

\section{Method}
\label{sec:method}
We present an image-immunization framework, termed Single-step Projection in Stochastic Neighborhoods (\textsc{SPIN}), for protecting images against unauthorized editing. To efficiently obtain a protection signal that reflects the perturbation's effect on the predicted edit, we use single-step projection to estimate the clean latent from an intermediate denoising state. We further optimize the perturbation over a local stochastic neighborhood to reduce dependence on a single inference path and promote generalization across editing instructions.

\subsection{Preliminaries and Problem Formulation}
\label{sec:preliminary}

Given a source image $x\in[0,1]^{H\times W\times 3}$ and an editing instruction $c$, a frozen editor $\mathcal G_\theta$ produces an edited image $y=\mathcal G_\theta(x,c;\omega)$, where $\omega$ denotes the sampling randomness. The editor iteratively denoises a latent state $z_t$ conditioned on $x$ and $c$, then decodes the final latent into $y$.

Image immunization seeks a bounded input perturbation that disrupts subsequent edits while preserving the source image's appearance. The protected image is $x_\delta=\Pi_{[0,1]}(x+\delta)$, where $\Pi_{[0,1]}$ clips pixel values to the valid range. For a given instruction $c$ and sampling randomness $\omega$, we formulate the protection objective as:
\begin{equation}
    \delta^\star = \underset{\|\delta\|_\infty\leq\varepsilon}{\arg\max}\;
    \mathcal L\left(
        \mathcal G_\theta(x_\delta,c;\omega),
        \mathcal G_\theta(x,c;\omega)
    \right),
    \label{eq:image_immunization}
\end{equation}
where $\varepsilon$ limits the perturbation magnitude to preserve input fidelity, and $\mathcal L$ measures the discrepancy between protected and clean editing outputs. Only $\delta$ is optimized, with all model parameters frozen. The perturbation remains fixed during evaluation under instructions and sampling randomness not used in optimization.

\subsection{Overall Framework}
\label{sec:overall_framework}

\begin{figure}[!htbp]
    \centering
    \includegraphics[width=0.95\linewidth,
                     trim=10pt 30pt 10pt 30pt,clip]{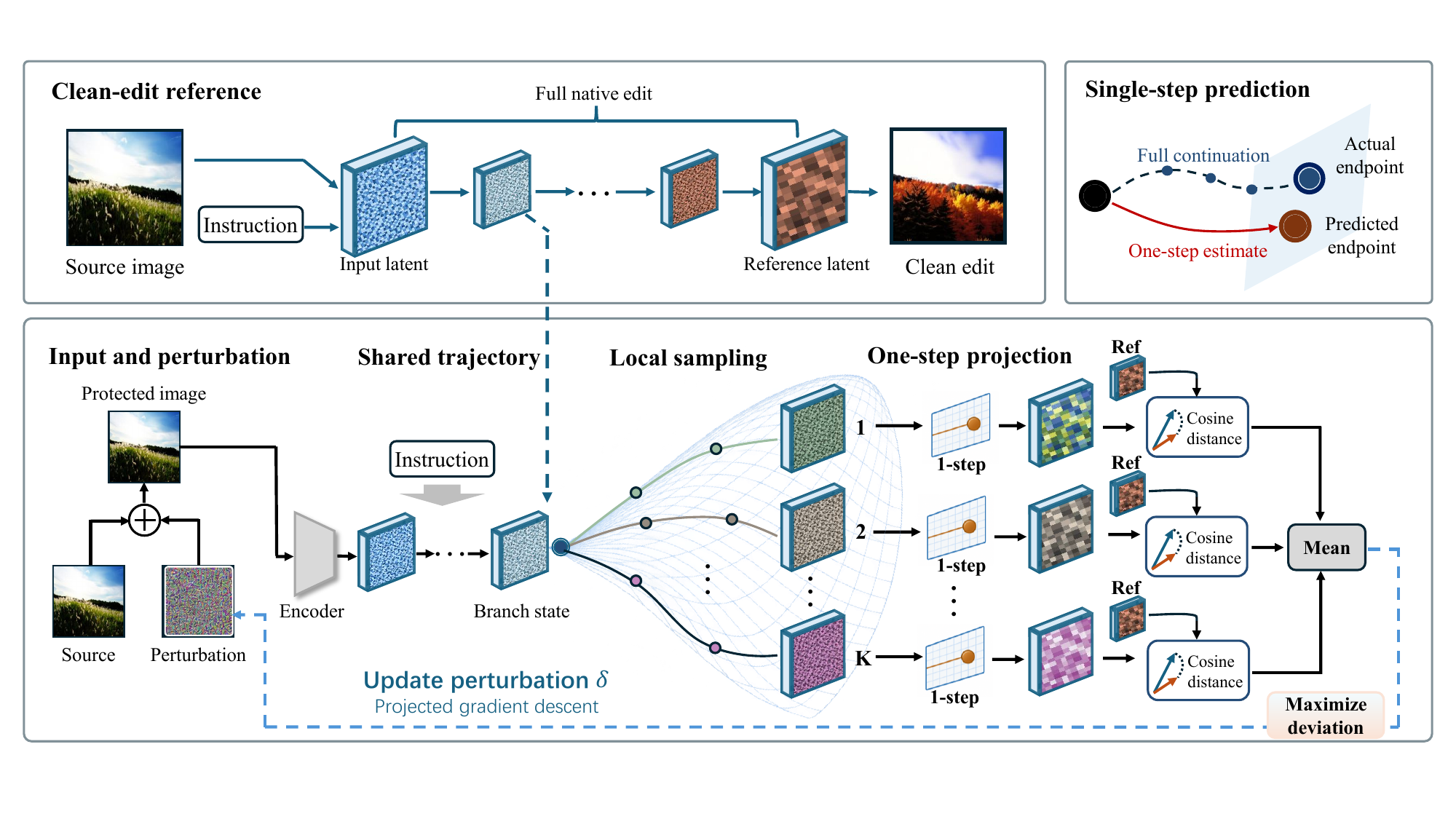}
    \caption{Overview of \textsc{SPIN}. From a shared native prefix, local trajectory sampling generates neighboring candidates and  one-step projection estimates their clean endpoints. These estimates guide perturbation optimization to drive editing outputs away from the fixed clean-edit reference.}
    \label{fig:spin_framework}
    \vspace{-10pt}
\end{figure}

Our goal is to construct a bounded input perturbation that disrupts the final editing output and remains effective under unseen editing instructions. To this end, \textsc{SPIN} combines two modules, as shown in Figure~\ref{fig:spin_framework}: (a) local trajectory sampling generates neighboring candidate states from a shared early state under a fixed construction instruction, providing feedback across multiple local continuations to reduce path dependence and support transfer to unseen instructions; (b) one-step projection estimates each candidate's clean latent, enabling output-oriented evaluation without full sampling rollouts. We aggregate the projected candidates' deviations from a fixed clean-edit reference to optimize the same input perturbation. The following subsections detail the projection, neighborhood construction, and perturbation optimization.

\subsection{Outcome-Oriented Single-Step Projection}
\label{sec:single_step_projection}

To evaluate local continuations without completing each sampling rollout, we use single-step projection to predict the clean latent from an intermediate state. For a noise-prediction model, consider $z_t=a_tz_0+b_t\epsilon$, with $\epsilon\sim\mathcal N(0,I)$. For a flow-matching model with a linear probability path, $z_t=(1-t)z_0+t\epsilon$. The corresponding projections are
\begin{equation}
\begin{aligned}
    \mathcal P_t^\epsilon(z_t;x_\delta,c) &= \frac{z_t-b_t\widehat\epsilon_\theta(z_t,t\mid E(x_\delta),c)}{a_t}, \\
    \mathcal P_t^v(z_t;x_\delta,c) &= z_t-t\widehat v_\theta(z_t,t\mid E(x_\delta),c).
\end{aligned}
\label{eq:single_step_projection}
\end{equation}
Here, $\widehat\epsilon_\theta$ and $\widehat v_\theta$ denote the editor's predictions, including guidance operations. Each projection uses the prediction at the candidate state and bypasses the remaining sampling steps.

For population-optimal conditional noise or velocity predictors trained with squared error, the unguided projection equals the conditional mean of the clean latent. In practice, we apply Equation~\ref{eq:single_step_projection} to guided predictions. To evaluate intermediate states, SVDD-PM~\citep{li2025SVDD} approximates their soft values using rewards computed on predicted clean samples. Following this look-ahead evaluation principle, we use the projected latent's deviation from a clean-edit reference as a differentiable protection surrogate.

Specifically, we obtain the reference latent $z_{\mathrm{ref}}$ by running the complete native editor once on the clean input under instruction $c$. The reference is held fixed throughout perturbation optimization, and the discrepancy is computed in the shared clean-latent space.

Projection placement determines the prefix cost and the endpoint information available for evaluation. SuperEdit~\citep{li2025superedit} observes that early sampling stages primarily affect global layout in InstructPix2Pix, while later stages refine local attributes and image details. This motivates using early projections to guide perturbation optimization toward disrupting the overall editing structure while keeping the sampling prefix short. The next subsection describes how we construct local candidates and aggregate their projected outcomes.

\subsection{Optimization over Stochastic Trajectory Neighborhoods}
\label{sec:trajectory_neighborhood}

Different editing instructions induce different denoising trajectories. Optimizing a protective perturbation along a single construction trajectory may cause it to overfit to that particular path. We therefore optimize over a stochastic neighborhood of local continuations, aiming to reduce trajectory-specific overfitting and improve transfer to unseen instructions.

Let $j$ index the branch position in sampling order, and let $\mathcal F_{0:j}$ denote the native sampling prefix. For the current protected input, we compute a shared intermediate state, generate $K$ candidates through a model-specific stochastic transition $\mathcal T_j$, and project each candidate:

\begin{equation}
\begin{gathered}
    z_j(\delta)=\mathcal F_{0:j}(x_\delta,c;\omega_{<j}), \qquad
    z_{j+1}^{(k)}=\mathcal T_j\bigl(z_j(\delta),\xi_k;x_\delta,c\bigr), \\
    \widetilde z_0^{(k)}=\mathcal P_{t_{j+1}}\bigl(z_{j+1}^{(k)};x_\delta,c\bigr).
\end{gathered}
\label{eq:branch_and_project}
\end{equation}

Here, $\xi_k\sim\mathcal N(0,I)$ are independently sampled branch noises for $k=1,\ldots,K$, and $t_{j+1}$ is the sampling time of the candidate states. Intuitively, these branches represent alternative ways the same early editing state could evolve. Recomputing the model prediction at each candidate yields a projected clean latent, allowing us to consider how the perturbation affects several possible editing outcomes without completing their denoising trajectories.

We maximize the expected cosine distance between candidate projections and the fixed clean-edit reference $z_{\mathrm{ref}}$, approximated by a finite candidate average:
\begin{equation}
\begin{aligned}
   J(\delta) &= \mathbb E_{\xi}\left[d_{\mathrm{cos}}\bigl(\widetilde z_0(\delta,\xi),z_{\mathrm{ref}}\bigr)\right], \\
   \widehat J_K(\delta) &= \frac{1}{K}\sum_{k=1}^{K}d_{\mathrm{cos}}\bigl(\widetilde z_0^{(k)},z_{\mathrm{ref}}\bigr).
\end{aligned}
\label{eq:mc_objective}
\end{equation}
The cosine distance is defined as $d_{\mathrm{cos}}(a,b)=1-\langle a,b\rangle/(\|a\|_2\|b\|_2)$. This objective combines feedback from multiple local continuations to update the same input perturbation, encouraging a larger average predicted deviation across the sampled neighborhoods.

The candidate count $K$ controls the number of samples used to approximate the expectation, while the stochastic transition scale determines the neighborhood's extent. We sample the candidate noises once and reuse them throughout optimization, yielding a fixed sample-average objective. The optimized perturbation is then held fixed and evaluated under unseen instructions using the complete native editing pipeline.

\subsection{Instantiating the Neighborhood in Different Editors}
\label{sec:editor_instantiations}

\textsc{SPIN} adapts to flow-matching and noise-prediction editors through the transition $\mathcal T_j$, while retaining the same projection objective and perturbation optimization. For flow-matching editors such as SD3, we adopt the stochastic transition used in Flow-grpo~\citep{liu2025flow} and Flow-TTRL~\citep{chen2026test}. Let $\Delta t_j=t_{j+1}-t_j<0$ and $v_j=\widehat v_\theta(z_j,t_j\mid E(x_\delta),c)$. We construct local candidates by adding Gaussian noise with a corresponding drift correction:
\begin{equation}
z_{j+1}^{(k)}=z_j+\left[v_j+\frac{g_j^2}{2t_j}\bigl(z_j+(1-t_j)v_j\bigr)\right]\Delta t_j+g_j\sqrt{-\Delta t_j}\,\xi_k, \qquad g_j=\lambda\sqrt{\frac{t_j}{1-t_j}}.
\label{eq:sd3_em_step}
\end{equation}
Here, $\xi_k{\sim}\mathcal N(0,I)$ and $\lambda\geq0$ controls the stochasticity. Setting $\lambda=0$ recovers the native FlowMatch Euler update. The drift correction follows from the velocity--score relation, with its derivation and endpoint handling provided in Appendix~\ref{app:editor_transitions}.

For noise-prediction editors such as InstructPix2Pix~\citep{brooks2023instructpix2pix}, the native Euler-Ancestral sampler already provides a stochastic transition. With clean-latent estimate $\widehat z_{0,j}=z_j-\sigma_j\widehat\epsilon_j$, we sample candidates as
\begin{equation}
z_{j+1}^{(k)}=z_j+\frac{z_j-\widehat z_{0,j}}{\sigma_j}\bigl(\sigma_{\mathrm{down},j}-\sigma_j\bigr)+\tau\sigma_{\mathrm{up},j}\xi_k.
\label{eq:temperature_transition}
\end{equation}
The coefficients $\sigma_{\mathrm{down},j}$ and $\sigma_{\mathrm{up},j}$ are determined by the native scheduler. The temperature $\tau\geq0$ controls the neighborhood scale, with $\tau=1$ recovering the native transition. In both instantiations, candidates share the sampling prefix and differ only in their branch noise. We recompute the model prediction at each candidate state and apply the corresponding single-step projection in Equation~\ref{eq:single_step_projection}. These transitions construct the optimization neighborhood.

\subsection{Perturbation Optimization}
\label{sec:optimization_cost}

We reparameterize each candidate transition by adding a Gaussian noise term to its conditional mean. Holding the sampled noise fixed during backpropagation allows us to differentiate the projection loss with respect to the input perturbation through the candidate prediction, local transition, shared prefix, and image encoder. Throughout optimization, model parameters are frozen, and the clean reference $z_{\mathrm{ref}}$ is treated as a constant.

We initialize the perturbation at zero and maximize Equation~\ref{eq:mc_objective} using projected gradient
ascent~\citep{madry2018pgd}.
Define the feasible set as
$\mathcal C(x)=
\{\delta:\|\delta\|_\infty\leq\varepsilon,\;
x+\delta\in[0,1]^{H\times W\times3}\}$.
Each update is computed by:
\vspace{-5pt}
\begin{equation}
    \delta^{(n+1)}
    =
    \Pi_{\mathcal C(x)}
    \left[
        \delta^{(n)}
        +
        \eta\,
        \operatorname{sign}
        \left(
            \left.
            \nabla_\delta\widehat J_K(\delta)
            \right|_{\delta=\delta^{(n)}}
        \right)
    \right],
    \label{eq:pgd_update}
\end{equation}
where $\eta$ is the step size.
At each iteration, we recompute the shared prefix for the current
protected input, generate the candidates, aggregate their
projection distances, and update the perturbation within the
feasible set.
After $M$ iterations, we return
$x_{\mathrm{adv}}=x+\delta^{(M)}$.
The perturbation budget, step size, and iteration count are
specified in the experimental setup.

\section{Experiments}
\vspace{-5pt}
\subsection{Setup and Evaluation Protocol}
\label{sec:experiment_setup}
\vspace{-5pt}

\paragraph{Benchmark Datasets.}
We construct a benchmark of 500 samples from Pico-Banana, covering landscapes, artworks, multi-object scenes, real-world photographs, news imagery, public figures, and indoor scenes. For text-conditioned protection methods, perturbations are optimized using the original prompt associated with each image. Each sample is evaluated with its original prompt and five semantically disjoint editing instructions, resulting in 3,000 image-instruction pairs. Additional dataset and instruction details are provided in Appendix~\ref{app:benchmark_details}.

\vspace{-2pt}
\paragraph{Editing Models and Evaluation Metrics.}
We evaluate InstructPix2Pix and Stable Diffusion 3 (SD3) as denoising-diffusion and flow-matching editors, respectively. We compare edits of clean and protected inputs using PSNR~\citep{wang2009mse}, SSIM~\citep{wang2004ssim}, FSIM~\citep{zhang2011fsim}, VIFP~\citep{sheikh2006vif}, and LPIPS~\citep{zhang2018lpips}. Lower values of the first four and higher LPIPS indicate stronger protection; the directions reverse for perturbation imperceptibility. Following SIFM~\citep{dong2025sifm}, we report Immunization Success Rate (ISR): the fraction of protected edits with instruction failure or substantial perceptual degradation, judged by consensus of Kimi K2.5~\citep{kimiteam2026kimi25} and Qwen3-VL~\citep{bai2025qwen3vl}. Higher ISR indicates stronger protection. See Appendices~\ref{app:evaluation_details} and~\ref{app:isr_protocol} for evaluation details.
\vspace{-2pt}

\vspace{-2pt}
\paragraph{Implementation Details.}
For each image, we optimize an additive perturbation $\delta$ under $\lVert\delta\rVert_\infty \leq 8/255$ while keeping the model parameters frozen. Starting from $\delta=0$, we perform 20 PGD steps with a step size of $1/255$, projecting the perturbation and clipping the protected image after each step. We use branch index $j=1$ for both editors and set $\tau=2$ for InstructPix2Pix. Sampling noise is fixed during optimization. Evaluation uses unseen seeds, with clean and protected inputs sharing the same editing instruction and sampling noise. Further implementation details are provided in Appendix~\ref{app:optimization_details}.
\vspace{-2pt}

\subsection{Main Results}
\label{sec:main_results}
\vspace{-2pt}
\paragraph{Comparison with State-of-the-Art Methods.}

We first compare \textsc{SPIN} with representative image-immunization methods, PhotoGuard-E/PhotoGuard-D~\citep{salman2023raising}, SDS~\citep{xue2024sds}, ED~\citep{chen2024editshield}, AdvDM~\citep{liang2023advdm}, MIST~\citep{liang2023mist}, SA~\citep{lo2024distraction}, DCT~\citep{bala2025dctshield}, and SIFM~\citep{dong2025sifm} on SD3 and InstructPix2Pix under the instructions used to construct the protected inputs. Table~\ref{tab:main_seen} shows that \textsc{SPIN} achieves the greatest output discrepancy and highest immunization success rate on both editors. Compared with objectives defined on intermediate features or noisy states, our projection objective targets deviations in predicted editing outcomes, aligning perturbation optimization more closely with final-edit disruption and thereby improving immunization performance.

\definecolor{lightgray}{gray}{0.93}        
\definecolor{lightblue}{RGB}{220, 235, 252}

\setlength{\aboverulesep}{0pt}
\setlength{\belowrulesep}{0pt}

\vspace{-5pt}
\begin{table}[!htbp]
    \centering
    \captionsetup{skip=0.5pt}
    \caption{Comparison with image-immunization baselines under original instructions. Lower PSNR/SSIM/VIFp/FSIM and higher LPIPS/ISR indicate better immunization performance.}
    \label{tab:main_seen}
    \begingroup
    \footnotesize
    \setlength{\tabcolsep}{1.6pt}
    \renewcommand{\arraystretch}{1.04}
    \resizebox{\linewidth}{!}{%
    \begin{tabular}{@{}l::cccccc@{\paneldash}cccccc@{}}
        \toprule
        \multirow{2}{*}{Method} & \multicolumn{6}{c@{\paneldash}}{\textbf{SD3}}
        & \multicolumn{6}{c@{}}{\textbf{InstructPix2Pix}} \\
        \cmidrule(lr){2-7}\cmidrule(lr){8-13}
        & PSNR $\downarrow$ & SSIM $\downarrow$ & VIFp $\downarrow$
        & FSIM $\downarrow$ & LPIPS $\uparrow$ & ISR $\uparrow$
        & PSNR $\downarrow$ & SSIM $\downarrow$ & VIFp $\downarrow$
        & FSIM $\downarrow$ & LPIPS $\uparrow$ & ISR $\uparrow$ \\
        \midrule
        PhotoGuard-E & 18.96 & 0.6689 & 0.2134 & 0.7890 & 0.3250 & 58\%
        & 13.93 & 0.4680 & \underline{0.1010} & \underline{0.6919} & \underline{0.5585} &  83\% \\
        PhotoGuard-D & 17.16 & 0.5668 & 0.1873 & 0.7682 & 0.3946 & 66\%
        & \underline{13.16} & \underline{0.4042} & 0.1087 & 0.7053 & 0.5094 & 83\% \\
        SDS & 17.92 & 0.5868 & 0.1878 & 0.7591 & 0.3798 & 64\%
        & 14.82 & 0.4704 & 0.1368 & 0.7330 & 0.4703 & 86\%\\
        ED & 19.88 & 0.6530 & 0.2419 & 0.8084 & 0.3176 & 60\%
        & 16.06 & 0.5098 & 0.1538 & 0.7652 & 0.4199 & 77\% \\
        AdvDM & \underline{13.54} & \underline{0.4519} & \underline{0.1059} & \underline{0.6600} & \underline{0.5391} & \underline{84\%}
        & 15.41 & 0.4839 & 0.1473 & 0.7634 & 0.4388 & 85\% \\
        MIST & 14.09 & 0.4664 & 0.1189 & 0.6808 & 0.5251 & 56\%
        & 15.44 & 0.4560 & 0.1219 & 0.7418 & 0.5060 & 86\%\\
        SA & 17.56 & 0.5875 & 0.1850 & 0.7577 & 0.4535 & 68\%
        & 16.69 & 0.5009 & 0.1411 & 0.7718 & 0.4970 & 86\%\\
        DCT & 17.25 & 0.5506 & 0.1671 & 0.7591 & 0.4206 & 70\%
        & 18.11 & 0.5765 & 0.1852 & 0.8107 & 0.4244 & 77\%\\
        SIFM & 15.92 & 0.4945 & 0.1468 & 0.7027 & 0.4988 & 82\%
        & 15.02 & 0.4818 & 0.1346 & 0.7568 & 0.5129 & \underline{91\%}\\
        \rowcolor{lightblue}
        \textsc{SPIN} & \textbf{11.20} & \textbf{0.3450} & \textbf{0.0677} & \textbf{0.5845} & \textbf{0.6730} & \textbf{90\%}
        & \textbf{10.93} & \textbf{0.3072} & \textbf{0.0649} & \textbf{0.6188} & \textbf{0.6974} & \textbf{94\%}\\
        \bottomrule
    \end{tabular}%
    }
    \endgroup
    \vspace{-6pt}
\end{table}
\FloatBarrier

\definecolor{lightgray}{gray}{0.93}        
\definecolor{lightblue}{RGB}{220, 235, 252} 
\vspace{-4pt}

\setlength{\aboverulesep}{0pt}
\setlength{\belowrulesep}{0pt}
\begin{table}[!htbp]
    \centering
    \captionsetup{skip=0.5pt}
    \caption{Generalization to five unseen instructions. Perturbations
    are optimized with the original instruction; Lower PSNR/SSIM/VIFp/FSIM and higher LPIPS/ISR indicate better immunization performance.}
    \label{tab:main_unseen}
    \begingroup
    \footnotesize
    \setlength{\tabcolsep}{1.6pt}
    \renewcommand{\arraystretch}{1.04}
    \resizebox{\linewidth}{!}{%
    \begin{tabular}{@{}l::cccccc@{\paneldash}cccccc@{}}
        \toprule
        \multirow{2}{*}{Method} & \multicolumn{6}{c@{\paneldash}}{\textbf{SD3}}
        & \multicolumn{6}{c@{}}{\textbf{InstructPix2Pix}} \\
        \cmidrule(lr){2-7}\cmidrule(lr){8-13}
        & PSNR $\downarrow$ & SSIM $\downarrow$ & VIFp $\downarrow$
        & FSIM $\downarrow$ & LPIPS $\uparrow$ & ISR $\uparrow$
        & PSNR $\downarrow$ & SSIM $\downarrow$ & VIFp $\downarrow$
        & FSIM $\downarrow$ & LPIPS $\uparrow$ & ISR $\uparrow$\\
        \midrule
        PhotoGuard-E & 19.80 & 0.6928 & 0.2397 & 0.8109 & 0.3007 & 67\%
        & \underline{14.37} & 0.4884 & \underline{0.1101} & \underline{0.7076} & \underline{0.5486} & 82\%\\
        PhotoGuard-D & 20.15 & 0.7039 & 0.2471 & 0.8146 & 0.2988 & 64\%
        & 16.76 & 0.4974 & 0.1565 & 0.7855 & 0.4493 & 81\%\\
        SDS & 18.79 & 0.6202 & 0.2147 & 0.7843 & 0.3549 & \underline{69\%}
        & 15.54 & 0.5068 & 0.1545 & 0.7536 & 0.4463 & 84\% \\
        ED & 20.64 & 0.6789 & 0.2673 & 0.8305 & 0.2954 & 50\%
        & 17.10 & 0.5482 & 0.1770 & 0.7951 & 0.3833 & 79\% \\
        AdvDM & \underline{14.36} & \underline{0.4775} & \underline{0.1182} & \underline{0.6834} & \underline{0.5216} & 68\%
        & 16.65 & 0.5176 & 0.1665 & 0.7833 & 0.4179 & 82\% \\
        MIST & 15.02 & 0.4942 & 0.1355 & 0.7063 & 0.5070 & 66\%
        & 16.00 & \underline{0.4745} & 0.1324 & 0.7565 & 0.4940 & 83\%\\
        SA & 18.73 & 0.5616 & 0.1762 & 0.7581 & 0.4662 & 58\%
        & 16.48 & 0.5231 & 0.1628 & 0.7692 & 0.4805 & 82\% \\
        DCT & 18.59 & 0.5886 & 0.1981 & 0.7679 & 0.4259 & 67\%
        & 18.40 & 0.6198 & 0.2275 & 0.8245 & 0.3772 & 80\% \\
        SIFM & 17.81 & 0.5408 & 0.1634 & 0.7403 & 0.4854 & \underline{69\%}
        & 15.62 & 0.5038 & 0.1557 & 0.7532 & 0.4889 & \underline{85\%}\\
        \rowcolor{lightblue}
        \textsc{SPIN} & \textbf{13.95} & \textbf{0.4396} & \textbf{0.1024} & \textbf{0.6486} & \textbf{0.5704} & \textbf{85\%}
        & \textbf{14.13} & \textbf{0.4208} & \textbf{0.1044} & \textbf{0.6990} & \textbf{0.5543} & \textbf{91\%} \\
        \bottomrule
    \end{tabular}%
    }
    \endgroup
    \vspace{-6pt}
\end{table}

\paragraph{Generalization to Unseen Editing Instructions.} We evaluate protection on semantically disjoint instructions excluded from perturbation optimization. Table~\ref{tab:main_unseen} shows that \textsc{SPIN} achieves ISR of 85\% on SD3 and 91\% on InstructPix2Pix, with a larger advantage over competing methods than under original instructions. This advantage is consistent with its optimization across multiple stochastic trajectories, which reduces dependence on a single editing path and helps protection remain effective when the instruction changes.
\vspace{-5pt}

\FloatBarrier
\vspace{-2pt}
\paragraph{Qualitative Comparison.}

\begin{figure}[!htbp]
    \centering

    \sbox{\ipTwoPLeft}{%
        \includegraphics[
            width=0.46\textwidth
        ]{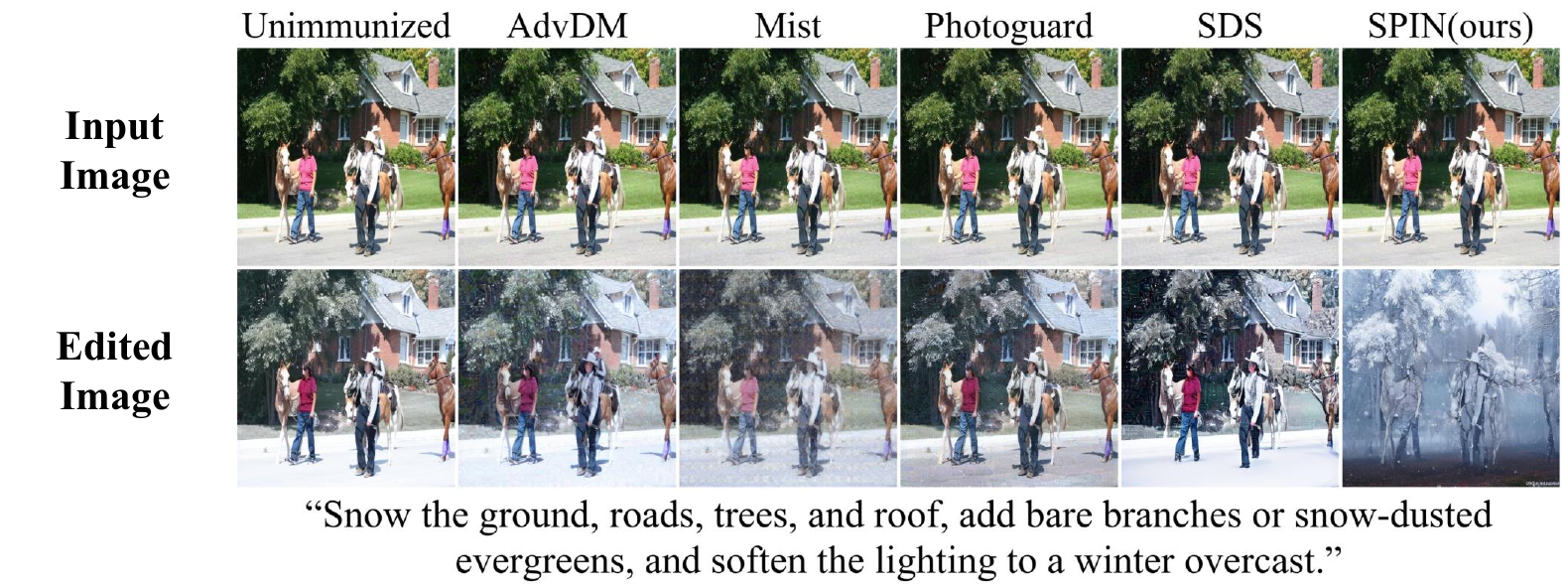}%
    }%

    \makebox[\textwidth][c]{%
        \usebox{\ipTwoPLeft}%
        \hspace{1mm}%
        \includegraphics[
            height=\ht\ipTwoPLeft,
            trim=100pt 0pt 0pt 0pt,
            clip
        ]{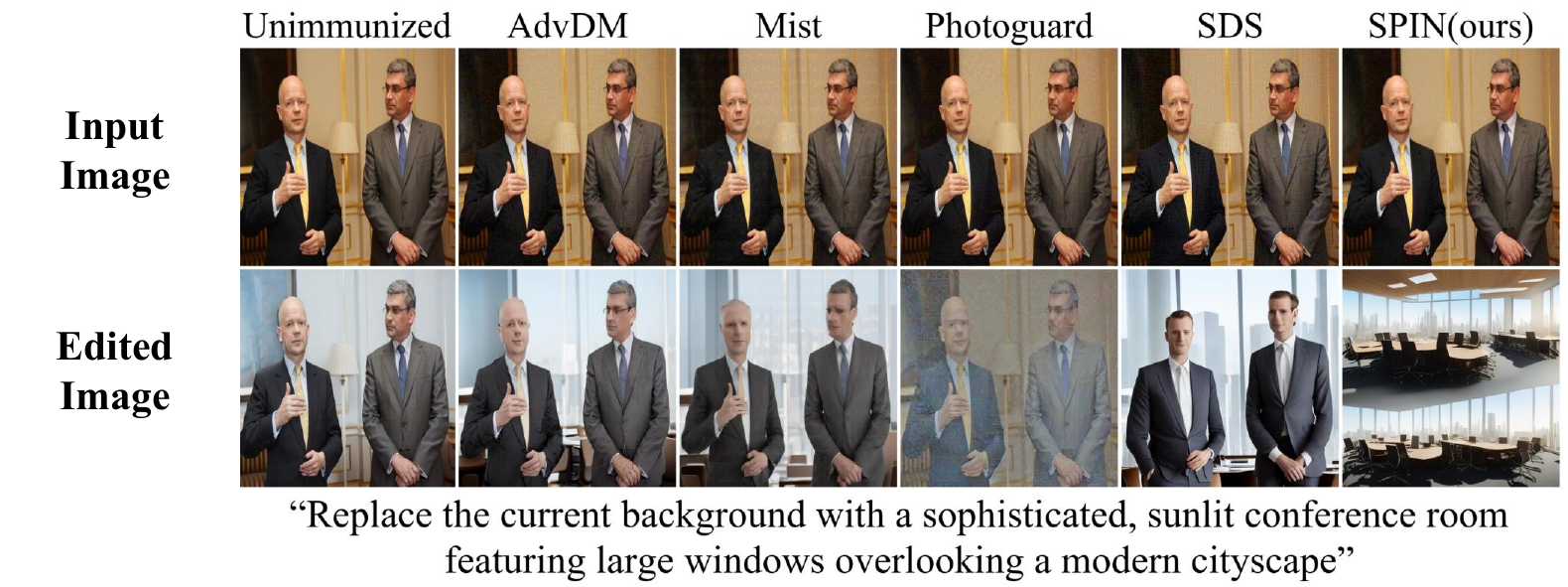}%
    }\\[1pt]

    \makebox[\textwidth][c]{%
        \includegraphics[
            width=0.46\textwidth
        ]{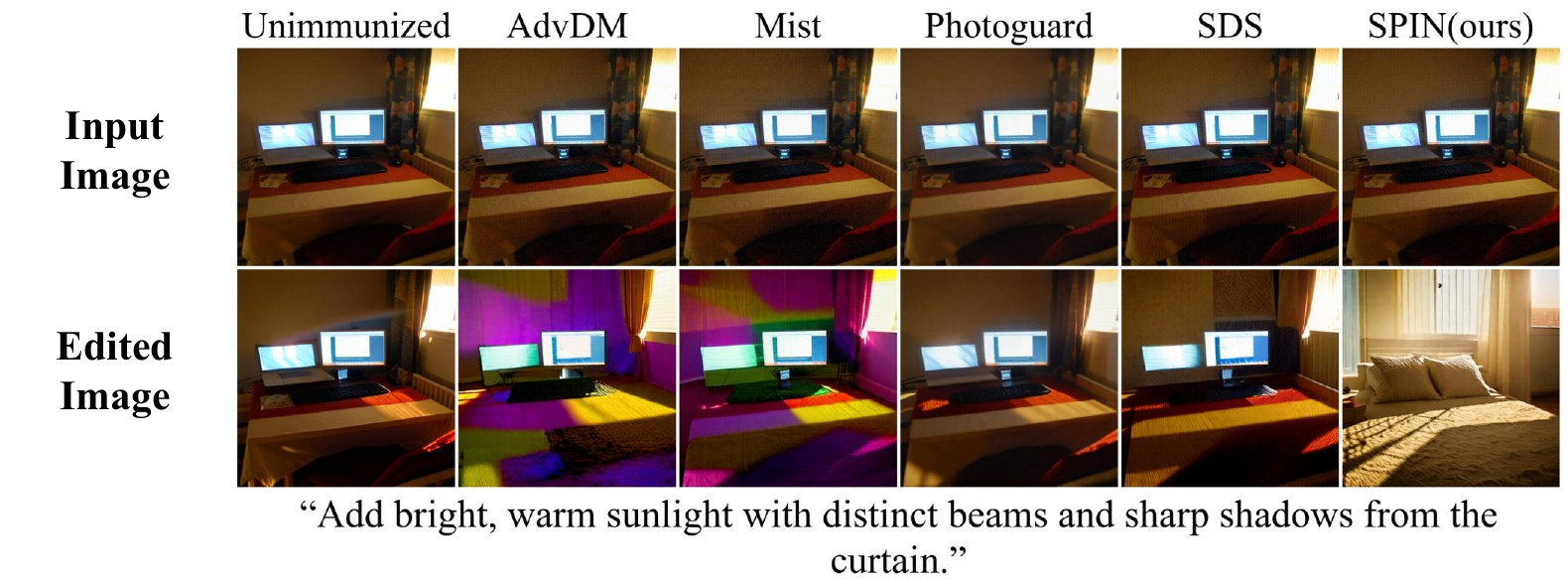}%
        \hspace{1mm}%
        \includegraphics[
            height=\ht\ipTwoPLeft,
            trim=100pt 0pt 0pt 0pt,
            clip
        ]{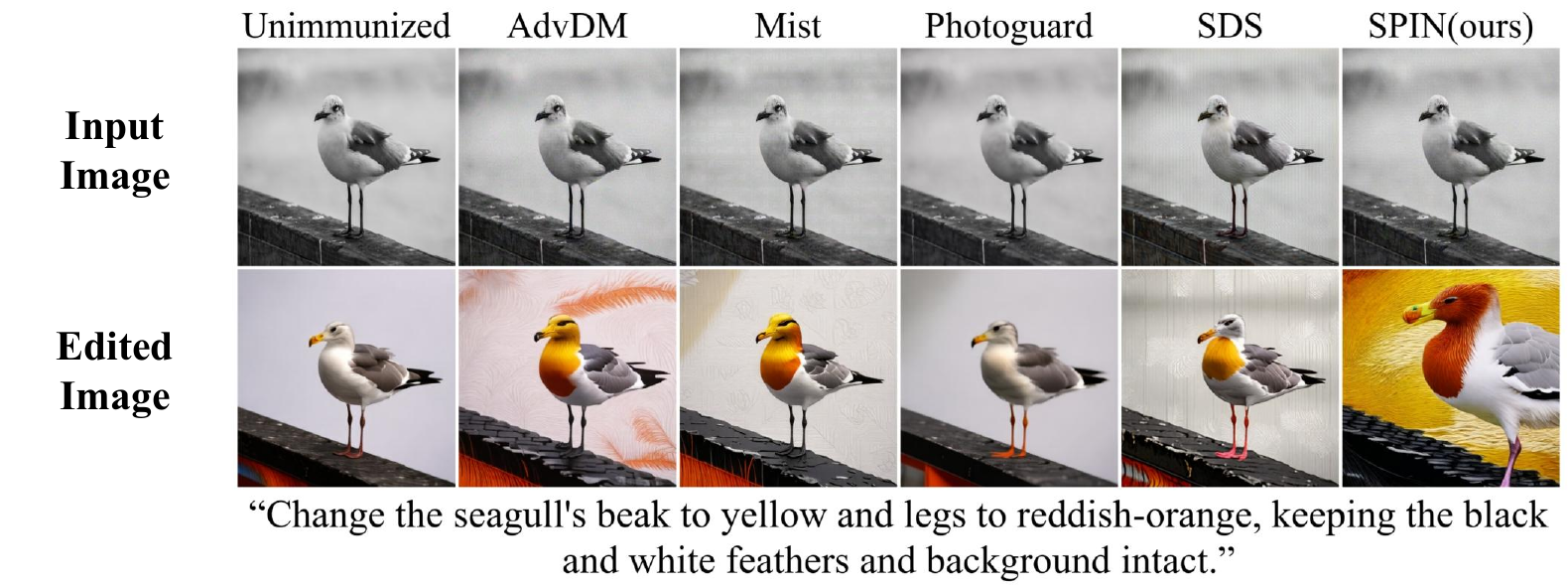}%
    }
    \begingroup
    \captionsetup{skip=2pt}
    \caption{Qualitative comparisons under seen editing instructions.
    The first and second rows show examples from InstructPix2Pix and SD3,
    respectively.}

    \label{fig:qualitative_comparison}
    \endgroup
    \vspace{-2pt}

    \sbox{\unseenLeft}{%
        \includegraphics[
            width=0.52\textwidth,
            trim=0pt 20pt 100pt 0pt,
            clip
        ]{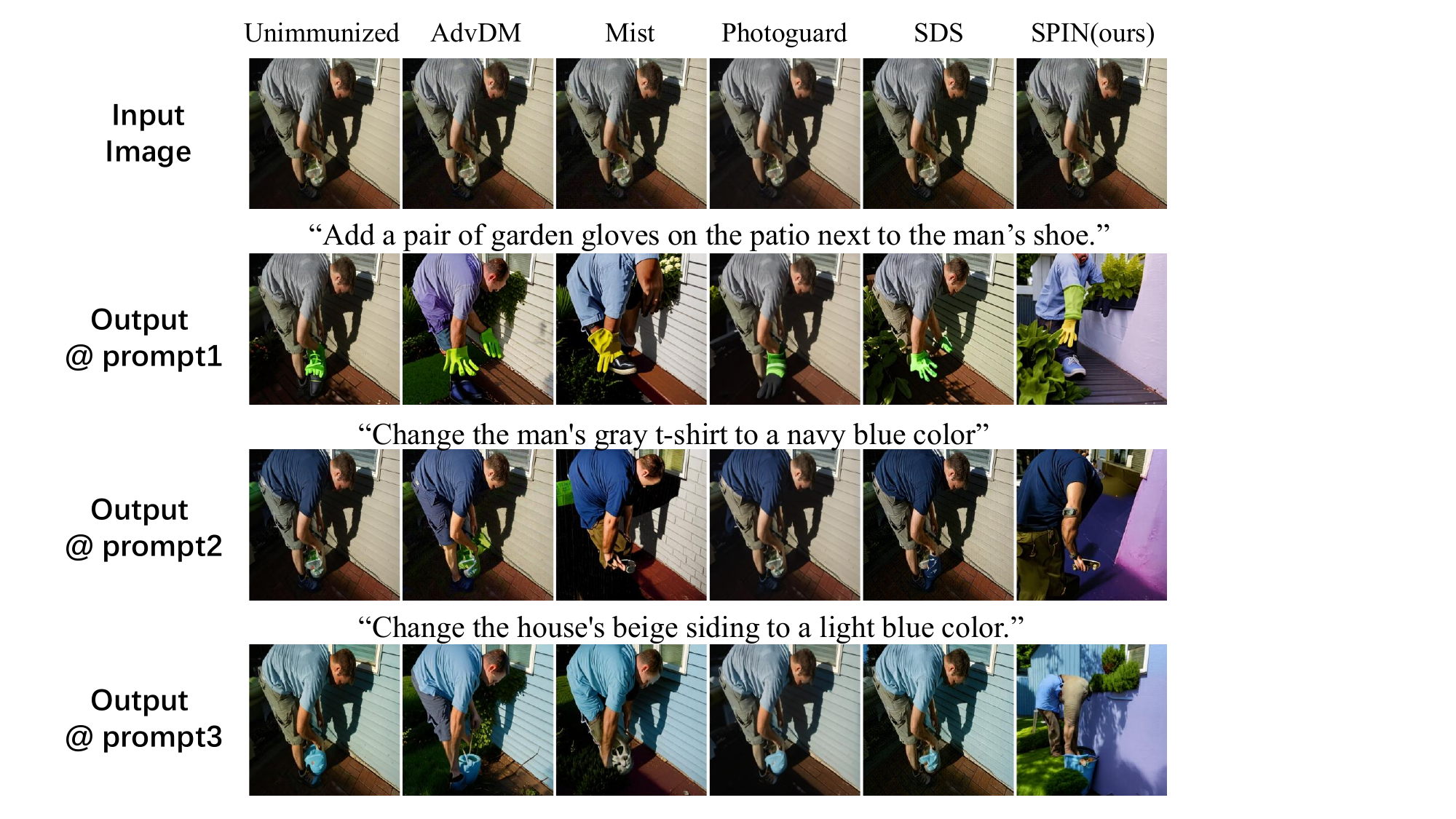}%
    }%
    \makebox[\textwidth][c]{%
        \usebox{\unseenLeft}%
        \hspace{-4mm}%
        \includegraphics[
            height=\ht\unseenLeft,
            trim=150pt 20pt 120pt 0pt,
            clip
        ]{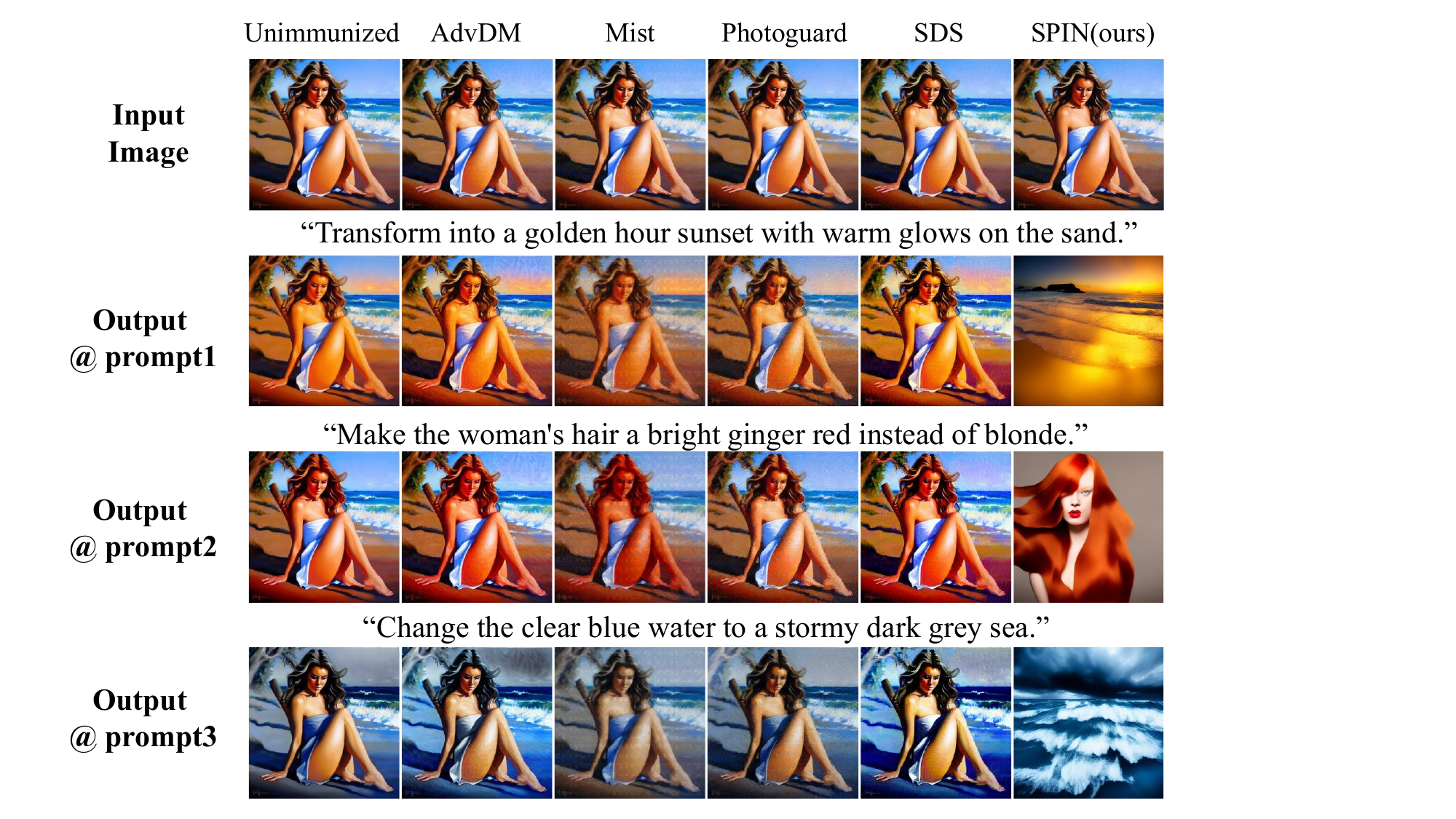}%
    }
    \begingroup
    \captionsetup{skip=2pt}
    \caption{Qualitative comparison under unseen editing instructions.
    The left and right examples use SD3 and IP2P, respectively, with editing
    instructions not used during protection optimization.}
    \label{fig:qualitative_unseen}
    \endgroup
    \vspace{-6pt}
\end{figure}

Figures~\ref{fig:qualitative_comparison} and~\ref{fig:qualitative_unseen} compare source images, protected inputs and editing results. Many baseline methods mainly introduce texture-like patterns or slight color shifts, while the requested edits remain largely intact. In contrast, \textsc{SPIN} causes substantial changes to the main subject or scene structure beyond the requested edits under both original and unseen instructions. More examples appear in Appendix~\ref{app:qualitative_unseen}.
\vspace{-2pt}

\vspace{-5pt}

\paragraph{Input Imperceptibility and Computational Efficiency.}

Protection quality also depends on input fidelity and perturbation-generation cost. We measure fidelity between clean and protected images separately from output discrepancy. All efficiency measurements are conducted on an NVIDIA GeForce RTX 5090 GPU, using the same input resolution and optimization budgets across methods. As shown in Table~\ref{tab:input_imperceptibility}, \textsc{SPIN} achieves better perturbation imperceptibility on most reported metrics and the highest immunization success rate, with a reasonable generation time per image.


\begin{table}[!htbp]
    \centering
    \captionsetup{skip=0.5pt}
    \caption{Input imperceptibility, immunization success, and
    time cost on SD3.}
    \label{tab:input_imperceptibility}
    \footnotesize
    \setlength{\tabcolsep}{2.5pt}
    \resizebox{0.80\linewidth}{!}{%
    \begin{tabular}{lccccccc}
        \toprule
        Method & PSNR $\uparrow$ & SSIM $\uparrow$ & VIFp $\uparrow$
        & FSIM $\uparrow$ & LPIPS $\downarrow$ & ISR $\uparrow$
        & Time/image (s) $\downarrow$ \\
        \midrule
        SDS          & 32.85 & 0.8439 & 0.5406 & 0.9700 & 0.1584 & 64\% & 10.7 \\
        PhotoGuard-E & 31.71 & 0.8635 & 0.5405 & 0.9801 & \textbf{0.0931} & 58\% &
        8.4 \\
        PhotoGuard-D & 33.24 & 0.8473 & 0.5412 & 0.9770 & 0.1835 & 66\% & 311.0 \\
        AdvDM        & 33.69 & 0.8511 & 0.5540 & 0.9758 & 0.1564 & \underline{84\%} & 14.9 \\
        MIST         & 33.23 & 0.8433 & 0.5496 & 0.9753 & 0.1611 & 56\% & 25.6 \\
        SIFM         & \underline{34.75} & \underline{0.8836} & \underline{0.5542} & \underline{0.9806} & 0.1279 & 82\% & 19.3 \\
        \rowcolor{lightblue}
        \textsc{SPIN} & \textbf{35.44} & \textbf{0.8950} & \textbf{0.6129} & \textbf{0.9812} & \underline{0.1187}
        & \textbf{90\%} & 16.0 \\
        \bottomrule
    \end{tabular}%
    }
    \vspace{-6pt}
\end{table}
\FloatBarrier

\subsection{Ablation Studies}
\label{sec:ablation_studies}

\paragraph{Effect of the Projection and Neighborhood Objectives.}
We measure latent discrepancy using the cosine distance between flattened latent representations. Table~\ref{tab:objective_ablation} reports this distance between the final latents of clean and protected edits after full denoising, alongside output PSNR and LPIPS. Under the same perturbation budget, one-step projection yields substantially larger latent cosine distances, lower PSNR, and higher LPIPS than random perturbations, demonstrating its effectiveness as a surrogate for disrupting final editing outputs. Compared with single-candidate projection, \textsc{SPIN} ($K=4$) further improves all three metrics on both editors under the same PGD budget, demonstrating the benefit of aggregating feedback across stochastic local continuations.

\begin{table}[!htbp]
    \centering
    \captionsetup{skip=0.5pt}
    \caption{Effect of one-step projection and stochastic neighborhood aggregation on protection efficacy. The deterministic SD3 variant uses the native deterministic transition, while stochastic variants use one or four candidates.}
    \label{tab:objective_ablation}
    \scriptsize
    \setlength{\tabcolsep}{3pt}
    \begin{tabular}{lccc}
        \toprule
        Variant & Latent cosine dist. $\uparrow$ & PSNR $\downarrow$
        & LPIPS $\uparrow$ \\
        \midrule
        \multicolumn{4}{l}{\textit{SD3}} \\
        Random perturbation                  & 0.0623 & 20.78 & 0.280 \\
        Deterministic one-step projection    & 0.2782 & 11.70 & 0.629 \\
        Stochastic one-step projection ($K=1$) & 0.2814 & 11.95 & 0.653 \\
        \textsc{SPIN} ($K=4$)                & \textbf{0.3141} & \textbf{11.20} & \textbf{0.673} \\
        \midrule
        \multicolumn{4}{l}{\textit{InstructPix2Pix}} \\
        Random perturbation                  & 0.1012 & 19.84 & 0.195 \\
        Stochastic one-step projection ($K=1$) & 0.7327 & 11.24 & 0.685 \\
        \textsc{SPIN} ($K=4$)                & \textbf{0.7637} & \textbf{10.93} & \textbf{0.697} \\
        \bottomrule
    \end{tabular}
    \vspace{-10pt}
\end{table}

\begin{figure}[!htbp]
\centering

\begin{subfigure}[c]{0.48\textwidth}
    \centering
    \captionsetup{skip=0pt} 
    \includegraphics[
        width=\linewidth,
        trim=0.6cm 0.42cm 0.88cm 0.2cm,
        clip
    ]{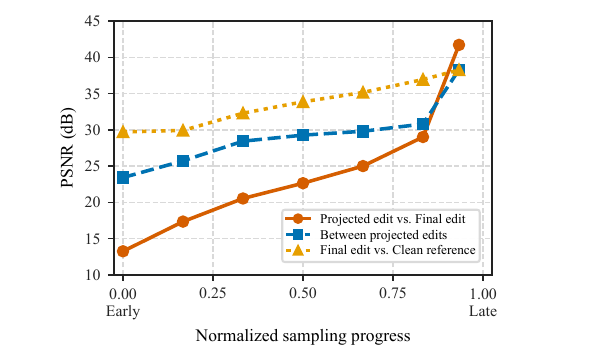}
    \vspace{-6pt} 
    \caption{Candidate diversity and deviation}
    \label{fig:branch_position_effect}
\end{subfigure}
\hspace{0.01\linewidth}%
\begin{subfigure}[c]{0.48\textwidth}
    \centering
    \captionsetup{skip=0pt} 

    \includegraphics[
        width=\linewidth,
        trim=0.6cm 0.42cm 0.88cm 0.2cm,
        clip
    ]{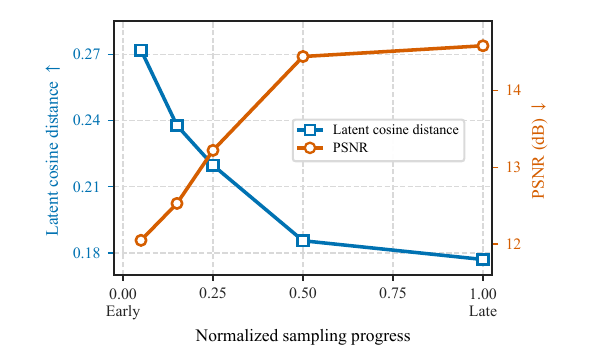}
    \vspace{-6pt} 
    \caption{Output deviation under guidance}
    \label{fig:projection_timing_effect}
\end{subfigure}

\caption{Effect of normalized sampling progress. (a) We report the PSNR
between a candidate's one-step projected edit and its fully denoised final
edit, between projected edits from different candidates, and between the
final edit and a fixed clean reference. Lower PSNR indicates a larger
difference. (b) Latent cosine distance and PSNR are both computed with
respect to the clean edit; higher latent cosine distance and lower PSNR
indicate a larger deviation.}
\label{fig:branch_position_ablation}
\vspace{-10pt}
\end{figure}

\paragraph{Effect of Branch Position and Guidance Timing.} We vary the branch position to examine candidate diversity and final-output deviation, and separately vary the timing of projection-guided latent updates that steer the trajectory away from the clean edit. Figure~\ref{fig:branch_position_ablation}(a) shows that later branching improves projection accuracy but reduces variation among candidate projections and their final deviations from the clean edit. Figure~\ref{fig:branch_position_ablation}(b) further shows that earlier latent updates produce larger final latent cosine distances and lower output PSNR, indicating stronger trajectory disruption. These observations motivate early branching and projection in \textsc{SPIN}.

\vspace{-20pt}

\paragraph{Effect of the Number of Candidate Trajectories.} We evaluate different numbers of candidate trajectories, $K\in\{1,2,4,8\}$. As shown in Table~\ref{tab:candidate_number_ablation}, increasing $K$ from 1 to 4 improves protection under both original and unseen instructions across the two editors. Increasing $K$ to 8 does not yield consistent performance improvement while substantially increasing generation time. We therefore use $K=4$ to balance protection effectiveness and computational cost.

\begin{table}[!htbp]
    \centering
    \captionsetup{skip=0.5pt}
    \caption{Effect of the number of candidate trajectories at a fixed
    editor-specific neighborhood scale.}
    \label{tab:candidate_number_ablation}
    \small
    \setlength{\tabcolsep}{4pt}
    \resizebox{\linewidth}{!}{%
    \begin{tabular}{lcccccc}
        \toprule
        Editor & $K$ & Seen PSNR $\downarrow$ & Seen LPIPS $\uparrow$
        & Unseen PSNR $\downarrow$ & Unseen LPIPS $\uparrow$
        & Time/image (s) $\downarrow$ \\
        \midrule
        SD3 & 1 & 11.95 & 0.6532 & 14.57 & 0.5702 & \textbf{10.51} \\
        SD3 & 2 & 10.99 & 0.6424 & 14.01 & 0.5718 & 12.65 \\
        SD3 & 4 & 11.20 & \textbf{0.6730} & \textbf{13.95} & 0.5704 & 16.01 \\
        SD3 & 8 & \textbf{10.94} & 0.6479 & 13.96 & \textbf{0.5828} & 27.11 \\
        \midrule
        InstructPix2Pix & 1 & 11.24 & 0.6848 & 15.51 & 0.5147 & \textbf{5.61} \\
        InstructPix2Pix & 2 & 10.88 & 0.6916 & 14.36 & 0.5316 & 7.40 \\
        InstructPix2Pix & 4 & 10.93 & 0.6974 & \textbf{14.13} & \textbf{0.5543} & 10.58 \\
        InstructPix2Pix & 8 & \textbf{10.41} & \textbf{0.7052} & 14.53 & 0.5531 & 20.97 \\
        \bottomrule
    \end{tabular}%
    }
    \vspace{-10pt}
\end{table}

\paragraph{Effect of the Perturbation Budget.} We vary the perturbation budget $\varepsilon$ while fixing the optimization budget and editor-specific settings. Figure~\ref{fig:epsilon_ablation} shows that larger perturbation budgets strengthen protection but reduce input fidelity. Increasing $\varepsilon$ beyond $8/255$ yields limited additional output discrepancy while introducing more visible input distortion, motivating our choice of $\varepsilon=8/255$.

\vspace{-30pt}
\newlength{\commonfigheight}
\setlength{\commonfigheight}{3.8cm}

\begin{figure}[!htbp]
    \centering

    \begin{minipage}[t]{0.30\linewidth}
    \centering
    \parbox[c][\commonfigheight][c]{\linewidth}{%
        \centering
        \includegraphics[
            width=\linewidth,
            height=\commonfigheight,
            keepaspectratio,
            trim=45pt 8pt 25pt 50pt,
            clip
        ]{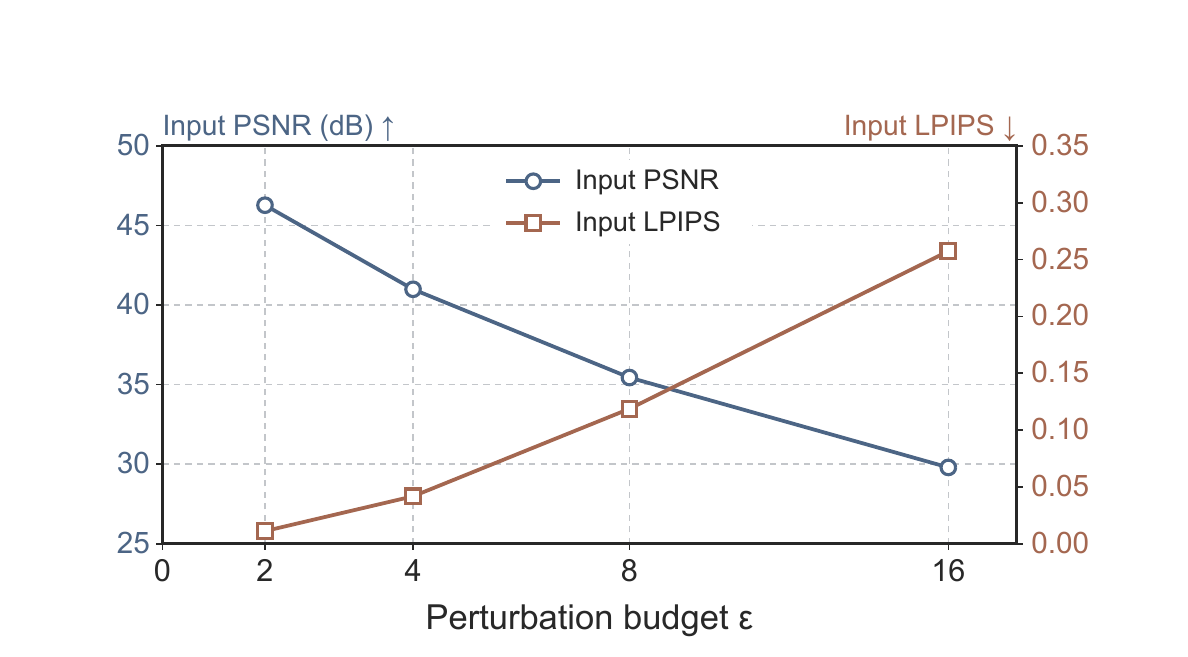}%

    }

    \par\vspace{-1.8em}
    \textbf{(a)} PSNR trade-off
\end{minipage}%
\hspace{0.01\linewidth}%
\begin{minipage}[t]{0.30\linewidth}
    \centering
    \parbox[c][\commonfigheight][c]{\linewidth}{%
        \centering
        \includegraphics[
            width=\linewidth,
            height=\commonfigheight,
            keepaspectratio,
            trim=45pt 8pt 25pt 50pt,
            clip
        ]{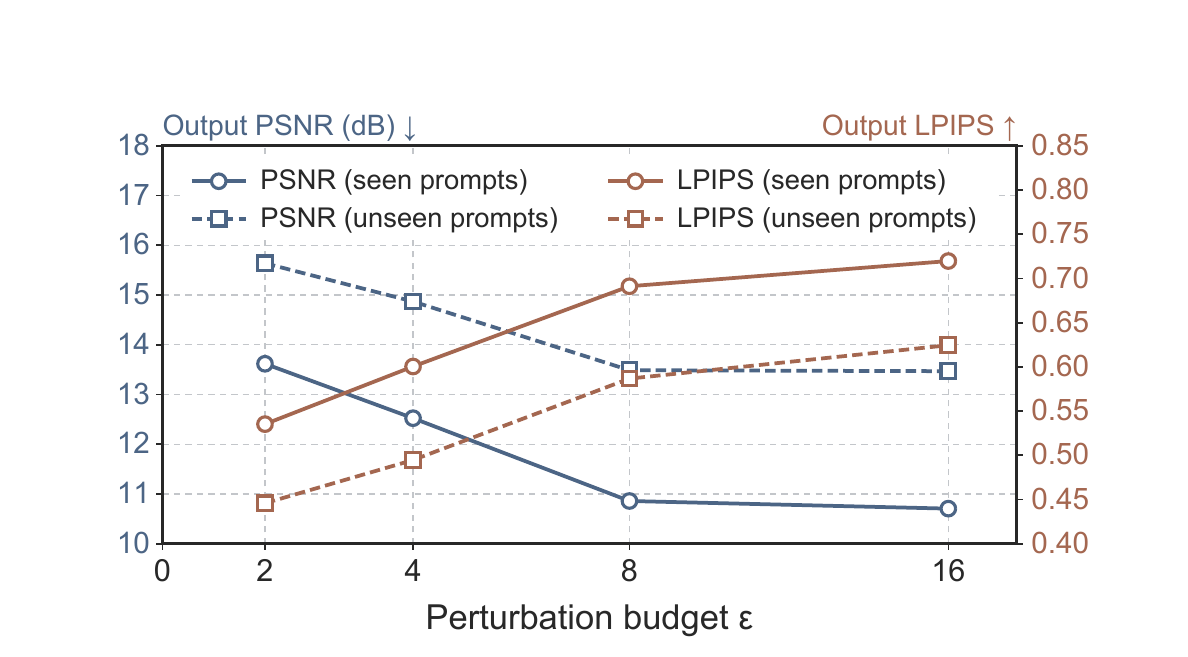}%
    }

    \par\vspace{-1.8em}
    \textbf{(b)} Output discrepancy
\end{minipage}%
\hspace{0.01\linewidth}%
\begin{minipage}[t]{0.35\linewidth}
    \centering
    \parbox[c][\commonfigheight][c]{\linewidth}{%
        \centering
        \includegraphics[
            width=\linewidth,
            height=\commonfigheight,
            keepaspectratio,
            trim=0pt 25pt 0pt 35pt,
            clip
        ]{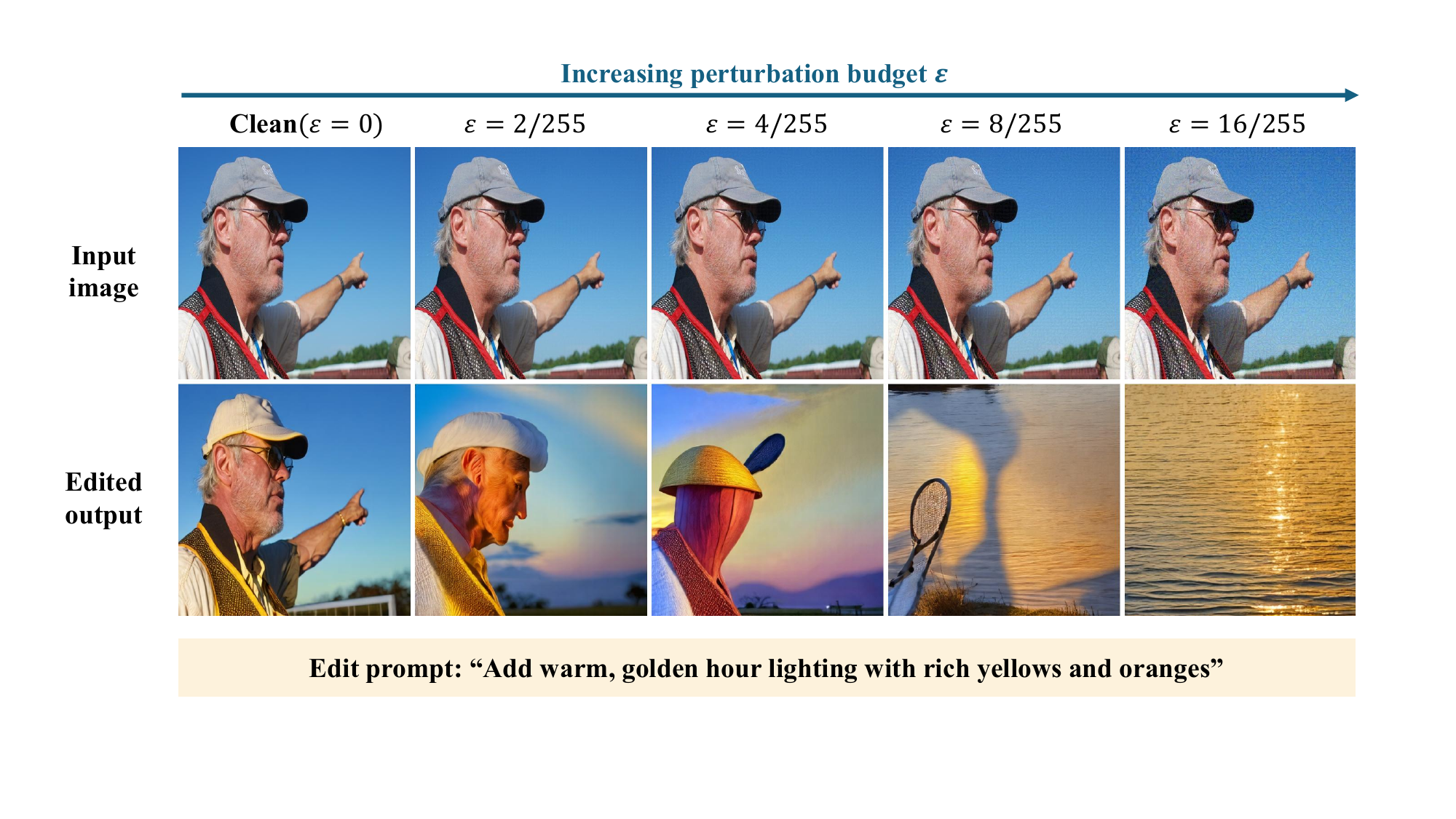}%
    }

    \par\vspace{-1.8em}
    \textbf{(c)} Qualitative results
\end{minipage}

    \caption{Effect of the perturbation budget $\varepsilon$: (a) input fidelity; (b) output discrepancy under seen and unseen instructions; (c) qualitative examples. Both (a) and (b) report PSNR and LPIPS, with budgets in units of $1/255$.}
    \label{fig:epsilon_ablation}
\end{figure}

\vspace{-10pt}

\section{Conclusion}
\vspace{-5pt}
We introduced \textsc{SPIN} to protect images against unauthorized editing by jointly considering the predicted outcomes of multiple local editing trajectories. This formulation makes output-oriented protection tractable without differentiating through the full denoising process. It also allows a protective perturbation to account for alternative continuations of an edit, rather than relying on a single construction path. Together, these ideas offer a practical direction for image immunization when future editing requests are unknown at the time of image release.
\FloatBarrier

\subsection*{AI use statement}

In this work, we used generative AI tools to assist with portions of the code implementation and to improve the clarity and readability of the manuscript. These tools helped draft selected code segments and refine wording, sentence structure, and the expression of technical content. The core research idea was developed by the authors without the use of generative AI. We also used Kimi K2.5 and Qwen3-VL as evaluators to compute the Immunization Success Rate (ISR), following the protocol described in Appendix D.
All AI-generated code was reviewed and checked by the authors. We also reviewed all AI-assisted text for accuracy and consistency with the intended meaning. We confirm the validity of the experimental results reported in this work. We take full responsibility for the final manuscript and accompanying code, including all claims, results, and content produced with AI assistance.

\subsection*{Ethics statement}

This work studies image immunization against unauthorized instruction-guided image editing. Although our method employs adversarial attacks, its objective is defensive: to help image owners protect their content by applying bounded perturbations before publication, reducing the risk of unwanted manipulation. The method operates on input images while keeping the editing models frozen. The same perturbation mechanism could also be used to disrupt legitimate editing workflows. We therefore encourage its use on images that users own or are authorized to protect. Our experiments use images from Pico-Banana, and the edited examples serve to evaluate protection effectiveness. They should not be interpreted as authentic depictions of people or events.

\subsection*{Reproducibility statement}

Section 3 describes the proposed method, and Section 4.1 summarizes the benchmark, editing models, metrics, and optimization settings. Appendices A and B provide additional evaluation details and the optimization algorithm. Evaluation uses seeds unseen during optimization, with each clean–protected pair sharing the same instruction and sampling noise to isolate the perturbation’s effect. The anonymized implementation is available at https://anonymous.4open.science/r/SPIN-778F/.

\bibliography{iclr2027_conference}
\bibliographystyle{iclr2027_conference}

\clearpage
\appendix
\addtocontents{toc}{\protect\setcounter{tocdepth}{2}}
\begingroup
\renewcommand{\contentsname}{Appendix Contents}
\tableofcontents
\endgroup
\bigskip
\section{Additional Experimental Details}
\label{app:experimental_details}

\subsection{Benchmark Construction}
\label{app:benchmark_details}

Our benchmark contains 500 samples drawn from Pico-Banana. The selected images
cover landscapes, artworks, multi-object scenes, real-world photographs, news
imagery, portraits of public figures, and indoor scenes. For protection methods
that require textual conditioning, the original prompt associated with each
image is used during perturbation optimization.

At evaluation time, each image is paired with its original prompt and five
additional editing instructions that are semantically disjoint from one
another. The additional instructions are used only for evaluation and are not
exposed during perturbation optimization. This separation tests whether the
protection transfers to editing requests that differ from the optimization
prompt. The resulting benchmark contains 3000
image--instruction pairs.

\subsection{Editing Models and Metric Computation}
\label{app:evaluation_details}

We evaluate protection efficacy using InstructPix2Pix and Stable Diffusion 3
(SD3), representing denoising-diffusion and flow-matching image editors,
respectively. All protection methods are evaluated using the same editor
configurations. For an input image $x$, its protected counterpart $x+\delta$,
an editing instruction $c$, and sampling noise determined by seed $s$, let

\begin{equation}
    \mathcal G_\theta(x,c;s)
    \quad\text{and}\quad
    \mathcal G_\theta(x+\delta,c;s)
\end{equation}
denote the editing outputs obtained from the clean and protected inputs. Both
outputs use the same instruction, inference configuration, and sampling noise.

Protection efficacy is evaluated by computing PSNR, SSIM, FSIM, VIFP, and
LPIPS between these two outputs. Lower PSNR, SSIM, FSIM, and VIFP and higher
LPIPS indicate stronger disruption of the editing result. Perturbation
imperceptibility is evaluated by applying the same metrics directly to $x$ and
$x+\delta$. In this setting, higher PSNR, SSIM, FSIM, and VIFP and lower LPIPS
indicate greater visual similarity between the clean and protected inputs.

\subsection{Protection Optimization and Randomness Control}
\label{app:optimization_details}

Given a clean image $x$, we optimize an additive perturbation $\delta$ while
keeping all parameters of the surrogate editor frozen. The perturbation is
initialized as $\delta^{(0)}=0$ and constrained by
\begin{equation}
    \lVert\delta\rVert_\infty \leq \frac{8}{255}.
\end{equation}
Optimization uses 20 projected-gradient steps with a step size of $1/255$.
After each update, the perturbation is projected onto the feasible
$\ell_\infty$ ball, and the protected image $x+\delta$ is clipped to the valid
pixel range. Algorithm~\ref{alg:local_trajectory_immunization} provides the
complete optimization procedure.

The surrogate editor uses fixed sampling noise across optimization iterations
to reduce stochastic variation in the objective. Evaluation uses inference
seeds that are not observed during perturbation optimization. Within each
evaluation pair, the clean and protected inputs share the same editing
instruction, inference seed, and sampling configuration, isolating the effect
of the input perturbation from unmatched generation randomness.

\section{Optimization Procedure}
\label{app:optimization_procedure}

Algorithm~\ref{alg:local_trajectory_immunization} gives the complete procedure
for constructing a protected input with local-trajectory one-step projection.

\begin{algorithm}[t]
\caption{Input Immunization via Local-Trajectory One-Step Projection}
\label{alg:local_trajectory_immunization}
\begin{algorithmic}[1]
\Require Frozen editor $\mathcal{G}=(E,\theta,\mathcal{S})$; source image $x$;
editing instruction $c$; branch index $j$; 
number of candidates $K$; perturbation budget $\varepsilon$;
step size $\alpha$; optimization iterations $M$
\Ensure Protected input $x_{\mathrm{adv}}$

\State Sample and fix native trajectory randomness $\omega_0$
\State $z_{\mathrm{ref}}\gets$ final latent of the full native edit
of $x$ under $c$ with noise $\omega_0$
\State Treat $z_{\mathrm{ref}}$ as a constant
\State Sample $\xi_k\overset{\mathrm{i.i.d.}}{\sim}\mathcal N(0,I)$
for $k=1,\ldots,K$ and keep them fixed
\State $\delta^{(0)}\gets 0$

\For{$n=0$ \textbf{to} $M-1$}
    \State $x_{\delta}\gets\Pi_{[0,1]}
        \left(x+\delta^{(n)}\right)$
      \State $z_j\gets
         \mathcal{F}_{0:j}(x_{\delta},c;\omega_{<j})$
         \Comment{Shared native prefix}
    \State $\widehat{J}_K\gets 0$

    \For{$k=1$ \textbf{to} $K$}
         \State $z_{j+1}^{(k)}\gets
               \mathcal{T}_{j}(z_j,\xi_k;x_{\delta},c)$
               \Comment{Local trajectory candidate}
         \State $\widetilde{z}_0^{(k)}\gets
            \mathcal{P}_{t_{j+1}}(z_{j+1}^{(k)};x_{\delta},c)$
            \Comment{One-step clean-latent projection}
        \State $\widehat{J}_K\gets\widehat{J}_K+
            \frac{1}{K}d_{\mathrm{cos}}
            \left(\widetilde{z}_0^{(k)},z_{\mathrm{ref}}\right)$
    \EndFor

    \State $g^{(n)}\gets
        \nabla_{\delta^{(n)}}\widehat{J}_K$
    \State $\bar{\delta}\gets
        \Pi_{[-\varepsilon,\varepsilon]}
        \left(\delta^{(n)}
        +\alpha\,\operatorname{sign}(g^{(n)})\right)$
    \State $\delta^{(n+1)}\gets
        \Pi_{[0,1]}(x+\bar{\delta})-x$
\EndFor

\State $x_{\mathrm{adv}}\gets x+\delta^{(M)}$
\State \Return $x_{\mathrm{adv}}$
\end{algorithmic}
\end{algorithm}

\section{Editor-Specific Transitions and Drift Correction}
\label{app:editor_transitions}

\subsection{Flow-Matching Transition Details}
\label{app:flow_transition}

For the flow-matching transition in Equation~\ref{eq:sd3_em_step}, the diffusion coefficient $g_j=\lambda\sqrt{t_j/(1-t_j)}$ controls both the drift correction and the random increment. Conditioned on the shared state $z_j$ and the editing conditions, each candidate follows
\begin{equation}
z_{j+1}^{(k)}\sim\mathcal N\!\left(\mu_j,g_j^2(-\Delta t_j)I\right), \qquad \mu_j=z_j+\left[v_j+\frac{g_j^2}{2t_j}\bigl(z_j+(1-t_j)v_j\bigr)\right]\Delta t_j.
\end{equation}
This construction follows the stochastic sampler used in Flow-TTRL~\citep{chen2026test}. If the scheduler includes $t_0=1$, we replace the denominator $1-t_0$ in $g_0$ with $1-t_1$, where $t_1<1$ is the first interior sampling time. This endpoint convention keeps the discrete noise coefficient finite. After branching, we evaluate the velocity separately at each candidate and compute
\begin{equation}
\widetilde z_0^{(k)}=z_{j+1}^{(k)}-t_{j+1}\widehat v_\theta\!\left(z_{j+1}^{(k)},t_{j+1}\mid E(x_\delta),c\right).
\end{equation}
Thus, candidate noise affects the projection through both the candidate state and its newly evaluated velocity.

\subsection{Derivation of the Drift Correction}
\label{app:drift_correction}

We restate the drift-correction derivation from Appendix~A of Flow-GRPO~\citep{liu2025flow} using our editing conditions and time convention. The construction follows the probability-flow ODE--SDE correspondence of \citet{song2021scorebased}. Consider the linear probability path $z_t=(1-t)z_0+t\epsilon$, where $\epsilon\sim\mathcal N(0,I)$ is independent of $z_0$ given the fixed conditioning information $h=(E(x_\delta),c)$. Let $p_t(z\mid h)$ denote the marginal density and $v^*(z,t\mid h)=\mathbb E[\epsilon-z_0\mid z_t=z,h]$ the exact velocity field. For $0<t<1$, the conditional density $p_t(z\mid z_0,h)$ is Gaussian with mean $(1-t)z_0$ and covariance $t^2I$. Taking the conditional expectation of its score over $z_0$ given $z_t=z$ and $h$ yields
\begin{equation}
s_t(z\mid h):=\nabla_z\log p_t(z\mid h)=-\frac{z-(1-t)\mathbb E[z_0\mid z_t=z,h]}{t^2}=-\frac{1}{t}\mathbb E[\epsilon\mid z_t=z,h].
\label{eq:app_score_identity}
\end{equation}
Taking conditional expectations in the identity $z_t+(1-t)(\epsilon-z_0)=\epsilon$ gives $\mathbb E[\epsilon\mid z_t=z,h]=z+(1-t)v^*(z,t\mid h)$. Consequently, the score can be expressed through the velocity:
\begin{equation}
s_t(z\mid h)=-\frac{z+(1-t)v^*(z,t\mid h)}{t}.
\label{eq:app_velocity_score}
\end{equation}

To verify the marginal consistency of the corrected stochastic process, introduce increasing sampling time $r=1-t$ and write $q_r(z\mid h)=p_{1-r}(z\mid h)$. The deterministic sampling flow has velocity $-v^*(z,1-r\mid h)$, so its density satisfies the continuity equation $\partial_r q_r=\nabla_z\cdot(v^*q_r)$. For a diffusion coefficient $g(t)$ depending only on time, consider
\begin{equation}
dZ_r=\left[-v^*(Z_r,t\mid h)+\frac{g(t)^2}{2}s_t(Z_r\mid h)\right]dr+g(t)\,dW_r, \qquad t=1-r.
\label{eq:app_forward_sampling_sde}
\end{equation}
Substituting $q_r$ into the corresponding Fokker--Planck equation gives
\begin{equation}
\begin{aligned}
-\nabla_z\cdot\left[\left(-v^*+\frac{g(t)^2}{2}s_t\right)q_r\right]+\frac{g(t)^2}{2}\Delta_z q_r
&=\nabla_z\cdot(v^*q_r)-\frac{g(t)^2}{2}\nabla_z\cdot(s_tq_r)+\frac{g(t)^2}{2}\Delta_z q_r\\
&=\nabla_z\cdot(v^*q_r),
\end{aligned}
\label{eq:app_fokker_planck}
\end{equation}
where the last equality uses $s_tq_r=\nabla_zq_r$. The score-dependent drift therefore cancels the diffusion contribution to the marginal density evolution. Under standard regularity and uniqueness conditions, the stochastic process and the deterministic flow share the same one-time marginals when initialized from the same marginal distribution~\citep{song2021scorebased}.

Returning to decreasing time $t$ and substituting Equation~\ref{eq:app_velocity_score}, we obtain
\begin{equation}
dZ_t=\left[v^*(Z_t,t\mid h)+\frac{g(t)^2}{2t}\bigl(Z_t+(1-t)v^*(Z_t,t\mid h)\bigr)\right]dt+g(t)\,d\overline W_t,
\label{eq:app_reverse_sde}
\end{equation}
where $\overline W_t$ denotes reverse-time Brownian motion, whose increment has covariance $(-dt)I$. An Euler--Maruyama step with $\Delta t_j<0$ gives Equation~\ref{eq:sd3_em_step} after replacing $v^*$ with the model prediction $v_j$. The marginal equivalence is a continuous-time result for exact velocity and score fields on intervals where the coefficients are well defined. In our implementation, guided predictions and finite sampling steps instantiate this construction as a local optimization transition.

\subsection{Noise-Prediction Transition Details}
\label{app:noise_prediction_transition}

For Euler-Ancestral sampling, $\sigma_j$ denotes the scheduler's noise level, and $\widehat\epsilon_j$ is the guided noise prediction evaluated with the scheduler's prescribed input scaling. The clean-latent estimate and update direction are
\begin{equation}
\widehat z_{0,j}=z_j-\sigma_j\widehat\epsilon_j, \qquad u_j=\frac{z_j-\widehat z_{0,j}}{\sigma_j}.
\end{equation}
The scheduler determines $\sigma_{\mathrm{down},j}$ and $\sigma_{\mathrm{up},j}$, which specify the deterministic update and the random increment in Equation~\ref{eq:temperature_transition}. Conditioned on the shared state, the resulting transition is
\begin{equation}
z_{j+1}^{(k)}\sim\mathcal N\!\left(z_j+u_j(\sigma_{\mathrm{down},j}-\sigma_j),\tau^2\sigma_{\mathrm{up},j}^2I\right).
\end{equation}
Changing $\tau$ preserves the conditional mean and scales the noise covariance, while $\tau=1$ recovers the native Euler-Ancestral transition. The temperature controls the neighborhood's extent independently of the candidate count $K$. After branching, we recompute the noise prediction at each candidate, using the scheduler's input scaling at the new noise level, and obtain
\begin{equation}
\widetilde z_0^{(k)}=z_{j+1}^{(k)}-\sigma_{j+1}\widehat\epsilon_{j+1}^{(k)}.
\end{equation}
For both editor families, the initial noise and all random increments in the shared prefix are fixed. Candidate noises are sampled once and reused throughout perturbation optimization, yielding a fixed sample-average objective. The optimized perturbation is subsequently evaluated with unseen instructions and fresh randomness through the complete native editing pipeline.

\section{ISR Evaluation Protocol}
\label{app:isr_protocol}

We follow the ISR protocol introduced by SIFM~\citep{dong2025sifm}. For each
evaluation sample, Kimi K2.5 and Qwen3-VL independently inspect the source
image, the edited output produced from the protected image, and the editing
instruction. Each evaluator returns binary judgments for two failure modes:
(1) semantic mismatch, where the output does not realize the intended edit,
and (2) substantial perceptual degradation, where artifacts, distortions, or
other quality defects make the output unusable. A model-level verdict is
positive when either failure mode is present.

Let $y_i^{(m)}\in\{0,1\}$ denote the verdict of evaluator $m$ for the $i$-th
editing attempt, where 1 indicates successful immunization. Following SIFM's
strict agreement rule, the aggregated verdict is
\begin{equation}
    y_i = y_i^{(\mathrm{Kimi})}\,y_i^{(\mathrm{Qwen})}.
\end{equation}
Thus, an attempt is counted as successfully immunized only when both evaluators
return positive verdicts; disagreements and jointly negative verdicts are
counted as failures. The evaluators are queried independently using the same
judging prompt and input presentation and do not observe each other's outputs.
For $N$ evaluated attempts, we compute
\begin{equation}
    \mathrm{ISR}=\frac{1}{N}\sum_{i=1}^{N} y_i.
\end{equation}
Higher ISR therefore indicates that a larger fraction of editing attempts are
jointly judged to violate the instruction or exhibit substantial perceptual
degradation.

\paragraph{System prompt.}
Both evaluators use the following system-prompt template. We replace
\texttt{\{instruction\}} with the editing instruction for each attempt and
provide the original, unprotected image as Image~1 and the edited result
obtained from its protected version as Image~2. The doubled braces in the
template become literal JSON braces after string formatting. Each evaluator
returns \texttt{YES} or \texttt{NO} for \texttt{semantic\_mismatch} and
\texttt{quality\_degradation}; a \texttt{YES} for either field yields a
positive model-level verdict.

\begin{lstlisting}[basicstyle=\ttfamily\footnotesize,columns=fullflexible,keepspaces=true,breaklines=true,breakatwhitespace=true,showstringspaces=false,frame=single,framerule=0.3pt,framesep=6pt,xleftmargin=6pt,xrightmargin=6pt]
JUDGE_PROMPT = """Edit instruction: "{instruction}"
Image 1: the original image (before editing).
Image 2: the edited result (after applying the edit instruction to a protected/immunized version of the original image).

Judge two criteria:

- semantic_mismatch: Does Image 2 FAIL to follow the edit instruction compared to Image 1? (YES/NO)
- quality_degradation: Does Image 2 have severe artifacts/noise/distortion making it unusable? (YES/NO)

Do not think or explain. Reply with ONLY this JSON, no other text: {{"semantic_mismatch":"YES or NO","quality_degradation":"YES or NO"}}"""
\end{lstlisting}

\clearpage
\section{Additional Qualitative Results under Unseen Instructions}
\label{app:qualitative_unseen}

\begin{figure}[!htbp]
    \centering
    \includegraphics[width=\linewidth]{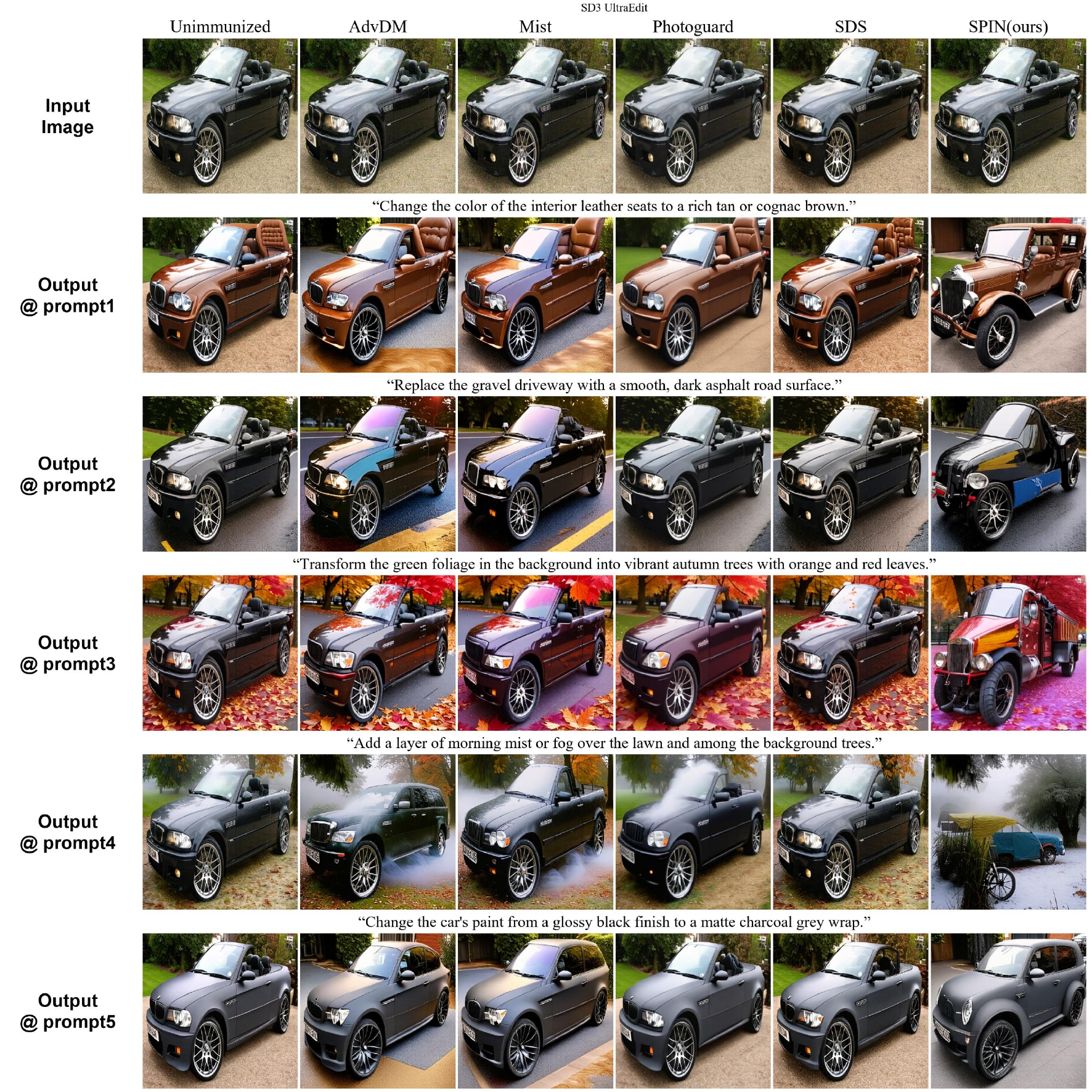}
    \caption{Unseen-instruction comparisons with SD3 on a car image.
    Columns show the unprotected input and the five protection methods.
    The top row shows the inputs; the remaining rows show edits under the
    five instructions printed above them. Each protected input is reused
    across all five instructions without re-optimization.}
    \label{fig:app_unseen_sd3_car}
\end{figure}

\begin{figure}[!htbp]
    \centering
    \includegraphics[width=\linewidth]{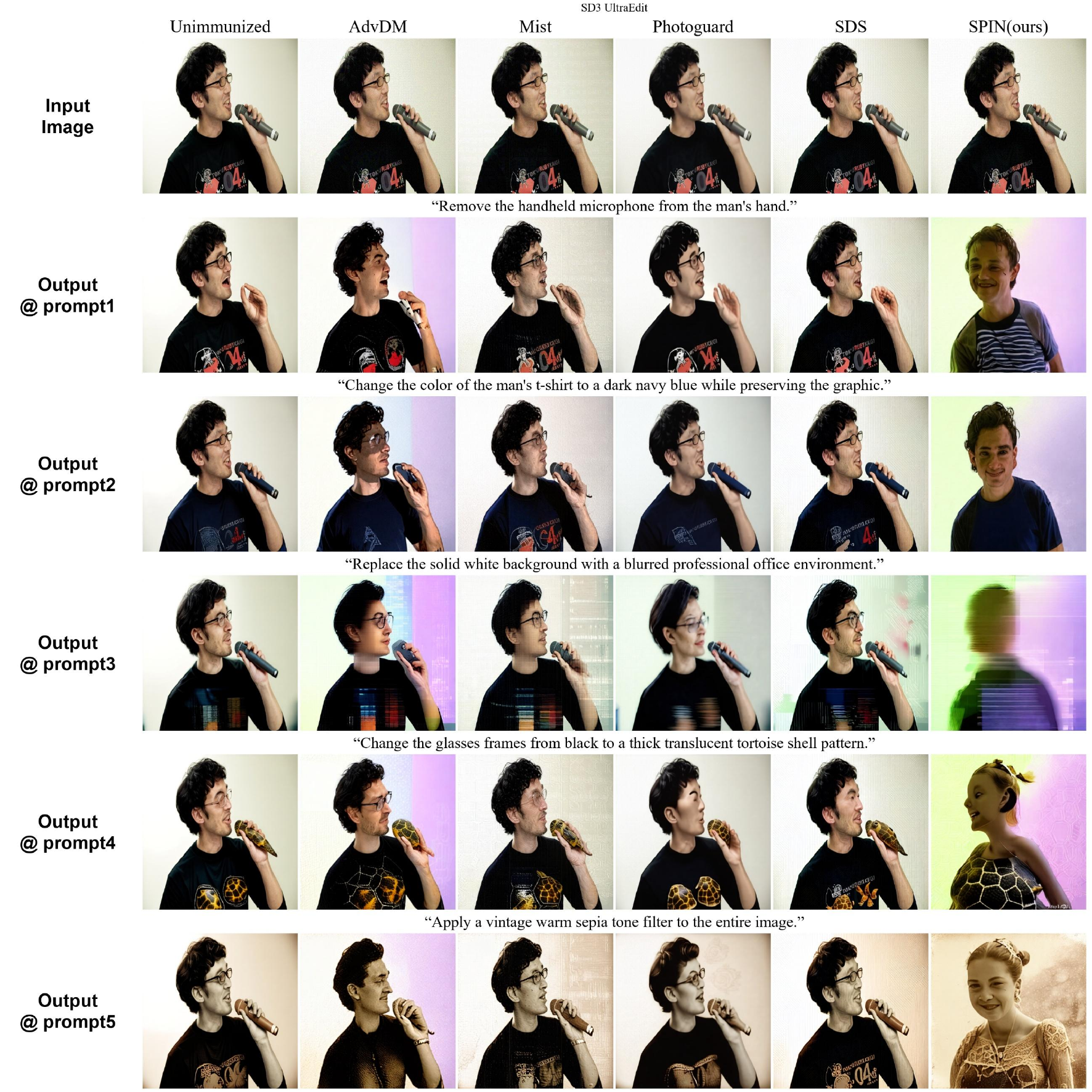}
    \caption{Unseen-instruction comparisons with SD3 on a portrait image.
    The top row shows the unprotected and protected inputs, and the remaining
    rows show outputs for five instructions excluded from protection
    optimization. The protected inputs remain fixed across instructions.
    The comparisons expose changes in subject appearance and image structure
    beyond the requested edits.}
    \label{fig:app_unseen_sd3_portrait}
\end{figure}

\begin{figure}[!htbp]
    \centering
    \includegraphics[width=\linewidth]{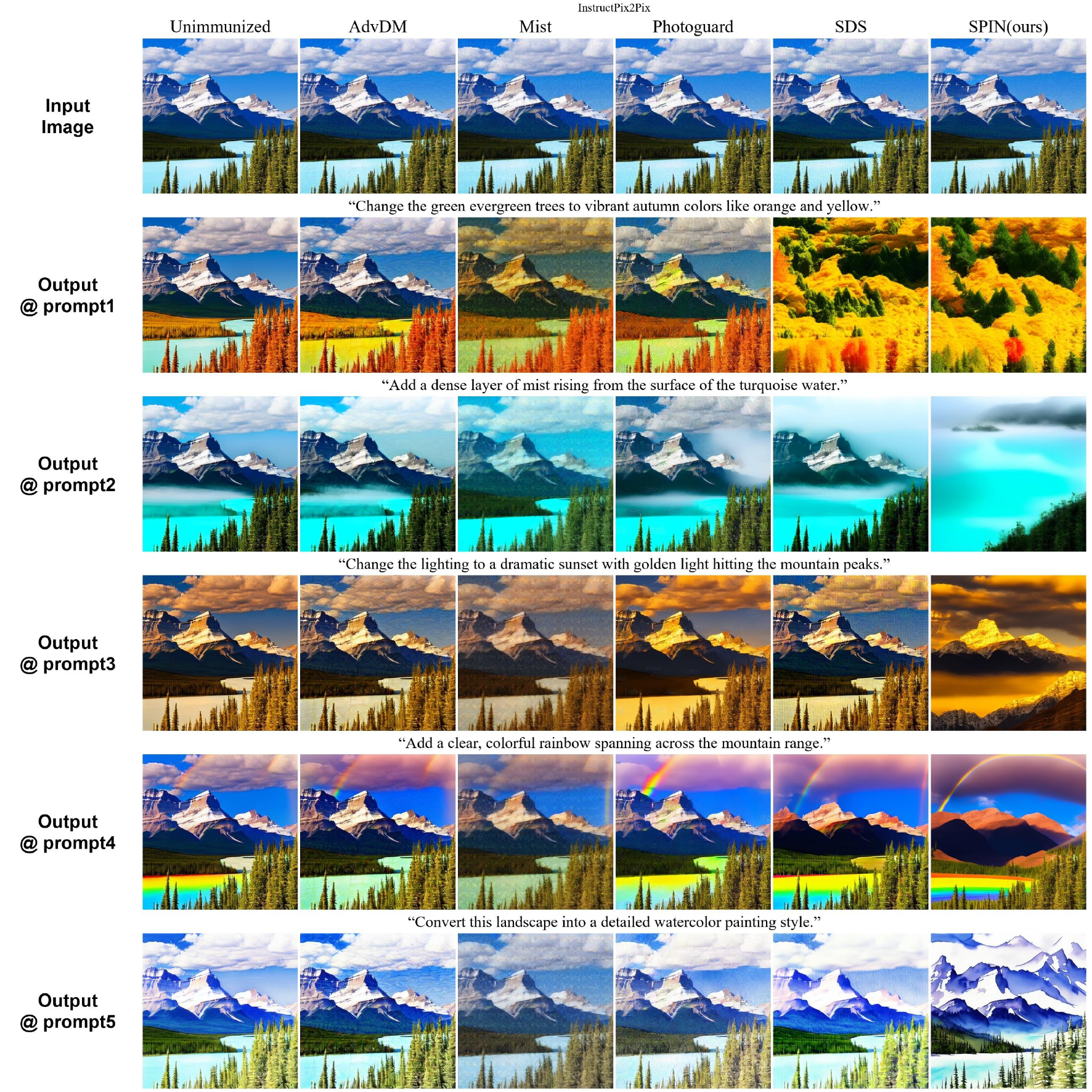}
    \caption{Unseen-instruction comparisons with InstructPix2Pix (IP2P)
    on a landscape image. The top row shows the unprotected and protected
    inputs; the remaining rows show edits under five unseen instructions
    using the same protected inputs. The instructions cover foliage color,
    mist, lighting, rainbow insertion, and watercolor style.}
    \label{fig:app_unseen_ip2p_landscape}
\end{figure}

\begin{figure}[!htbp]
    \centering
    \includegraphics[width=\linewidth]{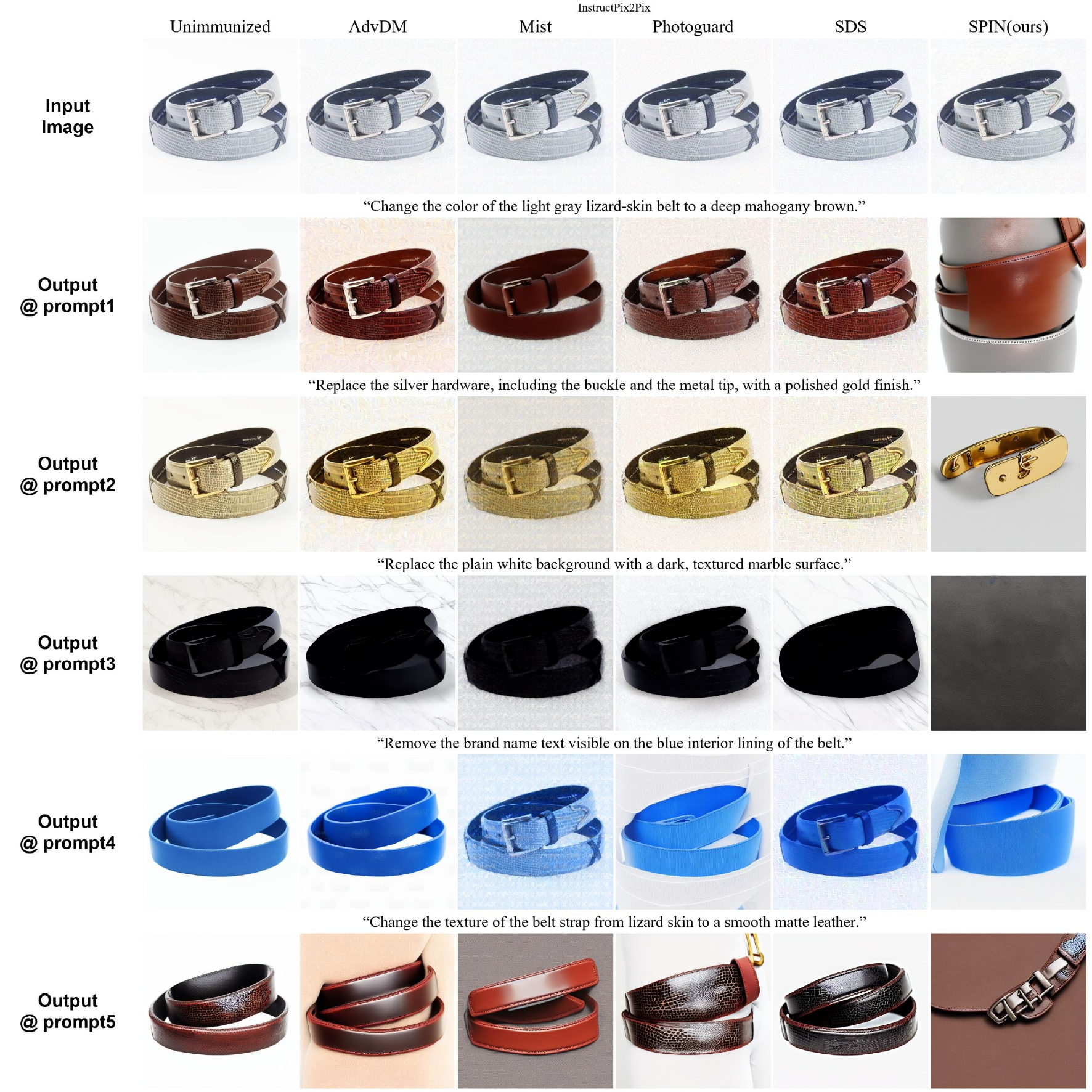}
    \caption{Unseen-instruction comparisons with InstructPix2Pix (IP2P)
    on a belt image. The top row shows the unprotected and protected inputs,
    followed by outputs for five instructions not used during protection
    optimization. The protected inputs remain fixed across instructions,
    which request changes to color, hardware, background, text, and texture.}
    \label{fig:app_unseen_ip2p_belt}
\end{figure}

Figures~\ref{fig:app_unseen_sd3_car} and~\ref{fig:app_unseen_sd3_portrait}
show additional SD3 examples, while
Figures~\ref{fig:app_unseen_ip2p_landscape} and~\ref{fig:app_unseen_ip2p_belt}
show InstructPix2Pix (IP2P) examples. Each source image is evaluated under
five editing instructions not used during protection optimization, with the
protected input held fixed across instructions. The comparisons include
unprotected inputs, AdvDM, Mist, PhotoGuard, SDS, and \textsc{SPIN}.
The examples illustrate changes beyond the requested edit: vehicle geometry
and portrait appearance change in the SD3 outputs, while scene layout and
object structure change in the IP2P outputs.

\clearpage

\clearpage

\clearpage

\end{document}